\documentclass[twoside,twocolumn,9pt]{article}

\usepackage{extsizes}
\usepackage[super,sort&compress,comma]{natbib} 
\usepackage[version=3]{mhchem}
\usepackage[left=1.5cm, right=1.5cm, top=1.785cm, bottom=2.0cm]{geometry}
\usepackage{balance}
\usepackage{mathptmx}
\usepackage{sectsty}
\usepackage{graphicx} 
\usepackage{lastpage}
\usepackage[format=plain,justification=justified,singlelinecheck=false,font={stretch=1.125,small,sf},labelfont=bf,labelsep=space]{caption}
\usepackage{float}
\usepackage{fancyhdr}
\usepackage{fnpos}
\usepackage[english]{babel}
\addto{\captionsenglish}{%
  
}
\usepackage{array}
\usepackage{droidsans}
\usepackage{charter}
\usepackage[T1]{fontenc}
\usepackage[usenames,dvipsnames]{xcolor}
\usepackage{setspace}
\usepackage[compact]{titlesec}
\usepackage{hyperref}

\usepackage{xr-hyper}   
\usepackage{amssymb}   
\usepackage{amsmath,amssymb}
\usepackage{physics}
\usepackage{booktabs}
\usepackage{pdfpages}

\usepackage{algorithm}
\usepackage{algpseudocode}  
\usepackage{setspace}     

\usepackage{epstopdf}

\definecolor{cream}{RGB}{222,217,201}

\begin{document}

\pagestyle{fancy}
\thispagestyle{plain}
\fancypagestyle{plain}{
\renewcommand{\headrulewidth}{0pt}
}

\makeFNbottom
\makeatletter
\renewcommand\LARGE{\@setfontsize\LARGE{15pt}{17}}
\renewcommand\Large{\@setfontsize\Large{12pt}{14}}
\renewcommand\large{\@setfontsize\large{10pt}{12}}
\renewcommand\footnotesize{\@setfontsize\footnotesize{7pt}{10}}
\makeatother

\renewcommand{\thefootnote}{\fnsymbol{footnote}}
\renewcommand\footnoterule{\vspace*{1pt}%
\color{cream}\hrule width 3.5in height 0.4pt \color{black}\vspace*{5pt}} 
\setcounter{secnumdepth}{5}

\makeatletter 
\renewcommand\@biblabel[1]{#1}            
\renewcommand\@makefntext[1]%
{\noindent\makebox[0pt][r]{\@thefnmark\,}#1}
\makeatother 
\renewcommand{\figurename}{\small{Fig.}~}
\sectionfont{\sffamily\Large}
\subsectionfont{\normalsize}
\subsubsectionfont{\bf}
\setstretch{1.125} 
\setlength{\skip\footins}{0.8cm}
\setlength{\footnotesep}{0.25cm}
\setlength{\jot}{10pt}
\titlespacing*{\section}{0pt}{4pt}{4pt}
\titlespacing*{\subsection}{0pt}{15pt}{1pt}

\fancyfoot{}
\fancyfoot[LO,RE]{\vspace{-7.1pt}\includegraphics[height=9pt]{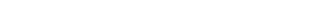}}
\fancyfoot[CO]{\vspace{-7.1pt}\hspace{13.2cm}\includegraphics{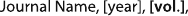}}
\fancyfoot[CE]{\vspace{-7.2pt}\hspace{-14.2cm}\includegraphics{head_foot/RF}}
\fancyfoot[RO]{\footnotesize{\sffamily{1--\pageref{LastPage} ~\textbar  \hspace{2pt}\thepage}}}
\fancyfoot[LE]{\footnotesize{\sffamily{\thepage~\textbar\hspace{3.45cm} 1--\pageref{LastPage}}}}
\fancyhead{}
\renewcommand{\headrulewidth}{0pt} 
\renewcommand{\footrulewidth}{0pt}
\setlength{\arrayrulewidth}{1pt}
\setlength{\columnsep}{6.5mm}
\setlength\bibsep{1pt}

\makeatletter 
\newlength{\figrulesep} 
\setlength{\figrulesep}{0.5\textfloatsep} 

\newcommand{\topfigrule}{\vspace*{-1pt}%
\noindent{\color{cream}\rule[-\figrulesep]{\columnwidth}{1.5pt}} }

\newcommand{\botfigrule}{\vspace*{-2pt}%
\noindent{\color{cream}\rule[\figrulesep]{\columnwidth}{1.5pt}} }

\newcommand{\dblfigrule}{\vspace*{-1pt}%
\noindent{\color{cream}\rule[-\figrulesep]{\textwidth}{1.5pt}} }

\makeatother

\twocolumn[
  \begin{@twocolumnfalse}
{\includegraphics[height=30pt]{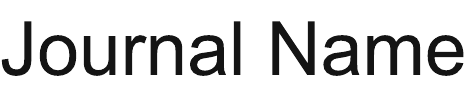}\hfill\raisebox{0pt}[0pt][0pt]{\includegraphics[height=55pt]{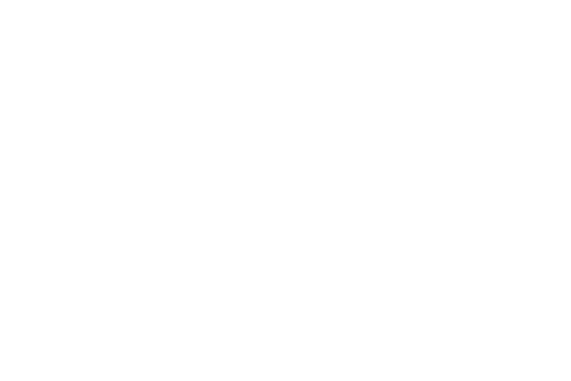}}\\[1ex]
\includegraphics[width=18.5cm]{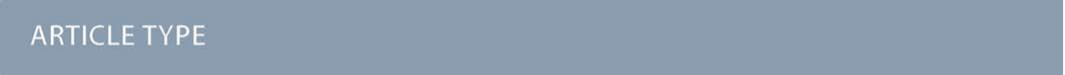}}\par
\vspace{1em}
\sffamily
\begin{tabular}{m{4.5cm} p{13.5cm} }

\includegraphics{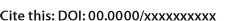} & \noindent\LARGE{\textbf{FlowMol3: Flow Matching for 3D De Novo Small-Molecule Generation}} \\
\vspace{0.3cm} & \vspace{0.3cm} \\

 & \noindent\large{Ian Dunn\textit{$^{a}$} \& David R. Koes\textit{$^{b*}$}} \\

\includegraphics{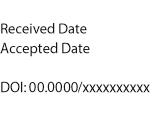} & \noindent\normalsize{

A generative model capable of sampling realistic molecules with desired properties could accelerate chemical discovery across a wide range of applications. Toward this goal, significant effort has focused on developing models that jointly sample molecular topology and 3D structure. We present FlowMol3, an open-source, multi-modal flow matching model that advances the state of the art for all-atom, small-molecule generation. Its substantial performance gains over previous FlowMol versions are achieved without changes to the graph neural network architecture or the underlying flow matching formulation. Instead, FlowMol3’s improvements arise from three architecture-agnostic techniques that incur negligible computational cost: self-conditioning, fake atoms, and train-time geometry distortion. FlowMol3 achieves nearly 100\% molecular validity for drug-like molecules with explicit hydrogens, more accurately reproduces the functional group composition and geometry of its training data, and does so with an order of magnitude fewer learnable parameters than comparable methods. We hypothesize that these techniques mitigate a general pathology affecting transport-based generative models, enabling detection and correction of distribution drift during inference. Our results highlight simple, transferable strategies for improving the stability and quality of diffusion- and flow-based molecular generative models.

} \\

\end{tabular}

 \end{@twocolumnfalse} \vspace{0.6cm}

  ]

\renewcommand*\rmdefault{bch}\normalfont\upshape
\rmfamily
\section*{}
\vspace{-1cm}


\footnotetext{\textit{$^{a}$Department of Computational and Systems Biology,
University of Pittsburgh, Pittsburgh, Pennsylvania 15260,
United States; Email: ian.dunn@pitt.edu}}
\footnotetext{\textit{$^{b}$Department of Computational and Systems Biology,
University of Pittsburgh, Pittsburgh, Pennsylvania 15260,
United States; Email: dkoes@pitt.edu}}


\footnotetext{$^{*}$~Corresponding Author}


\section{Introduction}

Deep generative models that can directly sample molecular structures with desired properties have the potential to accelerate chemical discovery by reducing or eliminating the need to engage in resource-intensive, screening-based  discovery paradigms. Moreover, generative models may improve chemical discovery by enabling multi-objective design of chemical matter. In pursuit of this idea, there has been recent interest in developing generative models for the design of small-molecule therapeutics,\cite{huang_dual_2024,guan_3d_2023,schneuing_structure-based_2023,peng_pocket2mol_2022,liu_generating_2022,torge_diffhopp_2023,igashov_equivariant_2024,dunn_accelerating_2023,cremer_flowr_2025, schneuing_multi-domain_2024} proteins,\cite{watson_novo_2023,bennett_atomically_2024,ingraham_illuminating_2023,yim_fast_2023, bose_se3-stochastic_2024} and materials \cite{zeni_mattergen_2024,miller_flowmm_2024, hoellmer_open_2025}.

In this work we focus on \textit{unconditional} generation of 3D, organic, small, drug-like molecules. Building models capable of accurate 3D molecule generation is a necessary precursor to accelerating chemical discovery. From a practical perspective, if a model cannot produce valid and synthetically accessible molecules, then it would be difficult to use it for real-world applications. Moreover, if a model struggles to produce reasonable molecules, it calls into question the ability of the generative model to learn even the most basic rules of chemistry. How can a generative model fulfill complex conditioning signals, such as the formation of hydrogen-bonding networks in binder design or selectivity and toxicity constraints for drug design, if that model cannot produce realistic molecules in the first place? Any improvements made to unconditional molecular generative models will have impacts in the design and performance of conditional generative models.


Deep generative models have delivered great advances in the \textit{de novo} design of molecules. Early attempts focused on generating either textual representations (SMILES strings) \cite{grisoni_bidirectional_2020,gomez-bombarelli_automatic_2018,dai_syntax-directed_2018} or 2D molecular graphs \cite{jin_junction_2019,liu_constrained_2019,shi_graphaf_2020,you_graph_2019}: molecular representations that exclude all information about 3D structure. Subsequent approaches were developed for 3D molecule generation using a variety of molecular representations and generative paradigms \cite{ragoza_learning_2020,ragoza_generating_2022,gebauer_symmetry-adapted_2020,luo_autoregressive_2021,satorras_en_2022}.

\begin{figure*}
    \centering
    \includegraphics[width=0.8\linewidth]{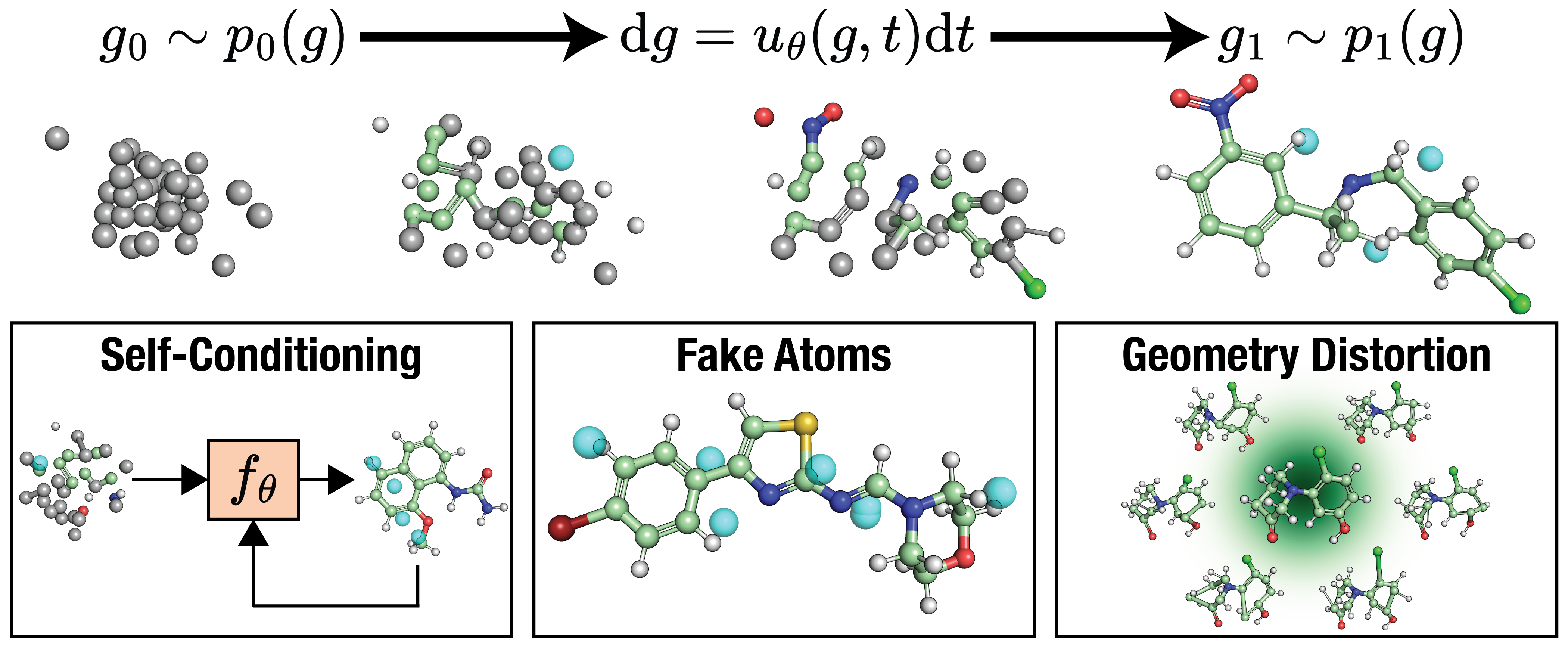}
    \caption{FlowMol3 Overview: FlowMol3 is a multi-modal flow matching model for unconditional \textit{de novo} molecule generation. Atomic coordinates are generated by sampling an ordinary differential equation learned via continuous flow matching. Atom types, charges, and bond orders are generated by simultaneous simulation of Continuous-Time Markov Chains (CTMCs) learned by discrete flow matching. We find three features to be critical for improved model performance: self-conditioning, fake atoms, and geometry distortion.}
    \label{fig:fm3-overview}
\end{figure*}

The emergence of diffusion models \cite{sohl-dickstein_deep_2015,ho_denoising_2020,song_score-based_2021} significantly changed this landscape,  following their success in computer vision. \citet{hoogeboom_equivariant_2022} applied the diffusion framework to an attributed point-cloud representation of molecules and showed a substantial improvement over existing methods. Subsequent works demonstrated that using discrete diffusion for categorical data, jointly modeling bond orders, and reparameterizing the denoising objective lead to further performance gains \cite{vignac_midi_2023,vignac_digress_2023,huang_learning_2023, peng_moldiff_2023, hua_mudiff_2024}. 


In parallel to the development of molecular diffusion models, flow matching emerged as a novel generative modeling framework \cite{lipman_flow_2023,tong_improving_2023,albergo_stochastic_2023,liu_flow_2022}. Flow matching generalizes diffusion models, offering simpler implementation and greater flexibility in model design. These advantages enabled Flow Matching to surpass the performance of diffusion models across various applications \cite{cremer_flowr_2025,schneuing_multi-domain_2024,bose_se3-stochastic_2024,yim_fast_2023,irwin_efficient_2024,dunn_exploring_2024,reidenbach_applications_2024,hoellmer_open_2025,miller_flowmm_2024}. Most relevant to our work here, the application of flow matching to \textit{de novo} molecule generation produced dramatic improvements in the capability of 3D molecular generative models \cite{irwin_efficient_2024,dunn_exploring_2024}.


Despite rapid progress in 3D \textit{de novo} molecule design, it remains apparent that generated molecules differ substantially from ``real'' molecules. Molecular generative models still produce invalid molecules, unrealistic geometries, and functional group compositions that deviate significantly from their training data \cite{nikitin_geom-drugs_2025,dunn_exploring_2024, buttenschoen_evaluation_2025}. 

More recent works have argued that relying on simplified, well-tested transformer-style architectures and scaling the size of the model will be essential to building powerful molecular generative models. This was the thesis of \citet{joshi_all-atom_2025} and \citet{reidenbach_applications_2024}. While scaling appears to have benefits, we argue there are other pathologies that cannot be remedied by architecture choice and scale.


Both diffusion and flow matching (which we refer to collectively as transport-based generative models) prescribe methods to transport samples between two distributions $q_{\text{source}}(x)$ and $q_{\text{target}}(x)$ by constructing a time-dependent process $p_t(x)$, which has the property that $p_0 = q_{\text{source}}$ and $p_1 = q_{\text{target}}$. Samples are drawn from the target distribution by iteratively sampling transition distributions $p_{t+dt|t}$, which are implicitly parameterized by a neural network. The learned process $p_t$ will only perfectly sample the target distribution in the limit of infinite data and a perfectly trained neural network. In reality, at every integration step the model imperfectly approximates $p_{t+dt|t}$, incurring drift from the desired marginal process that may accumulate through the sampling procedure.

We propose that transport-based generative models for \textit{de novo} molecule generation encounter difficulties primarily due to inference-time distribution drift that degrades the performance of the denoiser, and also an inability to correct distribution drift once it has occurred. Furthermore, we demonstrate several model features that we believe impart robustness to distribution drift and substantially improve molecule quality. These additional features are self-conditioning, modeling an extra "fake" atom type that enables the model to add or remove atoms from the system, and applying distortions to molecular structure on top of the interpolant. 

We present FlowMol3, a flow matching model for small molecule generation that substantially improves the state of the art in unconditional molecular generation. FlowMol3 is named so because it builds upon our previous iterations of this model\cite{dunn_exploring_2024,dunn_accelerating_2023}. The primary difference between FlowMol2 and FlowMol3 is the addition of features that, we argue, impart robustness to inference-time distribution drift. We show that the addition of these features alone dramatically alters model performance. Moreover, these changes do not significantly impact model size and introduce minimal computational overhead. These features in combination with a bespoke geometric graph neural network architecture enable FlowMol3 to achieve state-of-the-art performance while being substantially smaller than existing models.

\section{Model Design}

\subsection{Problem Setting}
We represent a 3D molecule with $N$ atoms as a fully-connected graph. Each atom is a node in the graph. Every atom has a position in space $X = \{x_i\}_{i=1}^N \in \mathbb{R}^{N \times 3}$, an atom type (in this case the atomic element) $A = \{a_i\}_{i=1}^N $, and a formal charge $C = \{c_i\}_{i=1}^N $. Additionally, every pair of atoms has a bond order $E = \{  e_{ij} \mid  i,j \in \{1,\dots,N\}, i \neq j \} $. Atom types, charges, and bond orders are categorical variables. For brevity, we denote a molecule by the symbol $g$, which can be thought of as a tuple of constituent data types $g = (X, A, C, E)$. 

We refer to these data types as ``modalities.'' We seek to build a flow matching model that can jointly sample the modalities that form our molecular graph $g$. FlowMol3 can thus be characterized as a ``multi-modal flow matching model.''

We sample atomic coordinates using Euclidean conditional flow matching \cite{tong_improving_2023,albergo_stochastic_2023,lipman_flow_2023, liu_flow_2022} and the remaining categorical modalities using discrete flow matching \cite{campbell_generative_2024, gat_discrete_2024}. In the following sections we briefly summarize the continuous and discrete flows used in FlowMol3. Then, we describe how we train one model to jointly sample interdependent modalities. 

\subsection{Flow Matching on Atomic Coordinates} \label{sec:fm-intro}

Flow matching \cite{tong_improving_2023,albergo_stochastic_2023,lipman_flow_2023, liu_flow_2022} is a method for learning to transport samples in a continuous space $x \in \mathbb{R}^d$ between two distributions $q_{\text{source}}(x)$ and $q_{\text{target}}(x)$ by constructing a time-dependent process $p_t(x)$ having the property that $p_0 = q_{\text{source}}$ and $p_1 = q_{\text{target}}$ and an ordinary differential equation $dx = u_t(x)dt$ which, when numerically integrated from initial positions $x_0 \sim p_{0}(x)$ up to time $t$, will produce samples distributed according to $p_t(x)$. 


After defining a time-independent conditioning variable $z \sim p(z)$, the marginal process $p_t$ is constructed as an expectation over conditional probability paths:


\begin{equation} \label{eq:marginal_path_def}
    p_t(x) = \mathbb{E}_{p(z)} \left[  p_t(x|z)  \right]
\end{equation}

The conditioning variable is generally taken to be either the final value $z=x_1$ or pairs of initial and final points $z=(x_0,x_1)$. A vector field $u_t(x)$ that produces the marginal process $p_t(x)$ can be approximated by regressing onto conditional vector fields:

\begin{equation} \label{eq:cfm-loss}
    \mathcal{L}_{CFM} = \mathbb{E}_{p(z), p_t(x|z), t\sim \mathcal{U}(0,1)} \norm{ u^\theta_t(x) - u_t(x|z) }^2
\end{equation}
where $u^\theta_t(x)$ is the learned vector field and $u_t(x|z)$ is a conditional vector field that produces $p_t(x|z)$. The remaining design choices for a continuous flow matching model are the choice of conditioning variable $z$, conditional probability paths $p_t(x|z)$, and conditional vector fields $u_t(x|z)$. Crucially, conditional probability paths are chosen so that they can be sampled in a simulation-free manner \footnote{Meaning that for any $t \in [0,1]$, we sample the conditional density $p_t(\cdot|z)$ in closed-form. To sample via simulation would entail sampling the prior $p_0(\cdot|z)$ and then performing numerical integration with the conditional velocity field $u_t(\cdot | z)$ up to time $t$, which makes training prohibitively expensive.} for any $t \in [0,1]$ and such that tractable closed-form expressions for the conditional vector-fields exist.

In FlowMol, the conditioning variable for atomic coordinate flows is paired initial and final atomic positions $z=(X_0,X_1)$. The distribution of our conditioning variable, $p(X_0,X_1)$, also known as the coupling distribution, is similar to the equivariant optimal transport coupling \cite{klein_equivariant_2023}. Essentially, we first obtain the $t=0$ atom coordinates as independent samples from a standard Gaussian $p_0(X)=\prod_{i=1}^N \mathcal{N}(x^i_0| 0, \mathbb{I})$. We then perform a rigid-body alignment and distance minimizing permutation of assignments between the prior positions $X_0$ and target positions $X_1$. This procedure can be seen in Algorithm \ref{alg:train_flowmol} and is discussed in detail in Section S3.


The conditional probability path for atomic coordinates is a Dirac density placed on a straight line connecting the terminal states.


\begin{equation} \label{eq:xcondpath}
    p_t(X|X_0,X_1) = \delta(X - (1 - t)X_0 - t X_1)
\end{equation}

This is equivalent to a deterministic interpolant \cite{albergo_stochastic_2023}:

\begin{equation} \label{eq:x-interpolant}
    X_t = (1 - t)X_0 + t X_1
\end{equation}

 The conditional probability path \eqref{eq:x-interpolant} is produced by the conditional vector field: \cite{dunn_mixed_2024,gat_discrete_2024,tong_improving_2023}

 \begin{equation} \label{eq:continuous-cond-vec-field}
     u_t(X|X_0,X_1) = \frac{1}{1-t}\left( X_1 - X_t \right)
 \end{equation}

 We apply ``endpoint parameterization'' as described previously\cite{dunn_mixed_2024}. Rather than letting the learned vector field $u^\theta_t(x)$ be the output of our neural network, we define our learned vector field as a function of the neural network output $\hat{X}_1(X_t)$:

 \begin{equation} \label{eq:learnvecfield}
     u^\theta_t(X_t) = \frac{1}{1-t}\left( \hat{X}_1(X_t) - X_t \right)
 \end{equation}

 By substituting the true conditional vector field \eqref{eq:continuous-cond-vec-field} and our chosen form of the approximate marginal vector field \eqref{eq:learnvecfield} into the conditional flow matching loss \eqref{eq:cfm-loss}, we obtain the endpoint flow matching loss \eqref{eq:denoiseloss}.

 \begin{equation} \label{eq:denoiseloss}
    \mathcal{L}_{EFM} = \mathbb{E}_{t,p_t(X_t|X_0,X_1),p(X_0,X_1)} 
    \left[ 
    w(t)\norm{\hat{X}_1(X_t) - X_1}^2
    \right]
 \end{equation}
where $w(t)=(1-t)^{-2}$ is a time-dependent weighting function. This approaches infinity as $t \to 1$. In practice, instead, we use a clamped weighting function $w(t) = \min(\max(0.005, \frac{t}{1-t}), 1.5)$.

\subsection{Discrete Flow Matching on Molecular Identity} 

\citet{campbell_generative_2024} and \citet{gat_discrete_2024} develop a flow matching method for sequences of discrete tokens, which we refer to as Discrete Flow Matching (DFM). In this section we will briefly summarize DFM using atom types as our sequences\footnote{The word ``sequence'' isn't quite correct, as it implies the existence of an inherent order. A more accurate conception of our data is as unordered or permutation-invariant sets of tokens. Nevertheless, the theory here applies seamlessly as DFM makes no explicit requirement that there be an order to sequences.} of interest, but this framework applies equivalently to the generation of atom formal charges and bond orders. 

In DFM, data are sequences of discrete tokens $A = \{a^i\}_{i=1}^N$ where each sequence element $a^i \in \{1,2,\dots,D\}$ (i.e., the atom type of a single atom) belongs to one of $D$ possible states. Each atom’s type evolves not by an ODE but by a \emph{continuous-time Markov chain (CTMC)}. A CTMC is the stochastic process where the sequence alternates between resting in its current state and jumping to another discrete state. 

DFM defines a CTMC on the interval $t\in[0,1]$ that transforms a sample from a simple prior distribution $A_0 \sim p_0(A) $ to a complex data distribution $A_1 \sim p_1(A)$. Jumps are governed by a probability velocity $u^i(j,A_t) \in \mathbb{R}$. This object describes the instantaneous flow of probability towards atom type $j$ for atom $i$, given the current sequence $A_t$.  This is analogous to the vector field in continuous flow matching. The marginal process $p_t(A)$ is simulated by iterative sampling of transition distributions that factorize over atoms:

\begin{equation} \label{eq:ctmctrans}
    p(A_{t+\Delta t}|A_t) = \prod_{i=1}^N p^i(a^i_{t+\Delta t}|A_t)
\end{equation}

Where $\Delta t$ is the integration step size , and $a^i_{t+\Delta t}$ is the atom type of the $i$-th atom at time $t+\Delta t$. The per-atom transition distributions are categorical with logits given by \eqref{eq:per-atom-transitions}.

\begin{equation} \label{eq:per-atom-transitions}
    p^i(a^i_{t+\Delta t}=j|A_t) = \delta \left(j, a_t^i\right) + u^i(j,A_t)\Delta t
\end{equation}
where $\delta$ is a Kronecker delta that returns one if its arguments are equal and zero otherwise. The primary task then becomes to find a tractable form of the probability velocity $u^i(j,A_t)$ that will generate our desired marginal process $p_t(A)$. As in continuous flow matching, the marginal process $p_t(A)$ is constructed as an expectation over conditional processes $p_t(A|A_0,A_1)$. Conditional probability paths factorize over each atom in the molecule:

\begin{equation} \label{eq:ctmc_cond_path} 
    p_t(A|A_0,A_1) = \prod_{i=1}^N p_t^i(a^i|A_0,A_1) 
\end{equation}

Where the per-atom conditional probability path takes the form:

\begin{equation} \label{eq:ctmc-condpath-peratom}
    p_t^i(a^i|A_0,A_1) = t\delta(A_1^i, a^i) + (1-t)\delta(M,a^i)
\end{equation}
where $M$ is an additional mask token added to the discrete states. The prior distribution is a Kronecker delta placed on a sequence of mask states, i.e., $p_0(A) = \prod_{i=1}^N \delta(M, a^i)$. The conditional path can be interpreted as follows: at time $t$, the $i$-th atom has probability $t$ of being in the masked state and probability $1 - t$ of being in its final state $a^i_1$. 


\citet{campbell_generative_2024} show that a marginal process constructed as such can be sampled with the following probability velocity for $j \neq a^i_t$:

\begin{equation} \label{eq:marginal_pvel}
    u^i(j,A_t) = \frac{1 + \eta t}{1 - t} p_{1|t}^{\theta}(a_1^i = j | A_t) \delta_M(a_t^i) + \eta (1 - \delta_M(a_t^i)) \delta_M(j)
\end{equation}

Where $\eta$ is the stochasticity parameter that, when increased, increases both the rate at which tokens are unmasked and remasked. To ensure that \eqref{eq:ctmctrans} defines a valid categorical distribution we set $u^i(a_t^i,A_t)= -\sum_{j\neq a_i^t}u^i(j,A_t)$. The only ``unknown'' quantity in the marginal probability velocity is a probability denoiser $p_{1|t}^{\theta}(a_1^i | A_t)$: the distribution of final states for atom $i$ given the current sequence $A_t$.  We train a neural network to approximate this distribution by minimizing the negative log-likelihood or, in practice, a standard cross-entropy loss. Note that this loss is applied only to atoms that are in the masked state at time $t$.

\begin{equation} \label{eq:dfm-loss}
    \mathcal{L}_{CE} = \mathbb{E}_{t,p(A_0,A_1),p_t(A_t|A_0,A_1)} 
    \left[ 
    -\sum_{i=1}^N
    a_1^i \log p_{1|t}^{\theta}(a_1^i | A_t) \delta_{M}(a^i_t)
    \right]
\end{equation}

Substituting our choice of $u^i(j,A_t)$ \eqref{eq:marginal_pvel} into the transition distribution \eqref{eq:ctmctrans} yields sampling dynamics that can be described simply. If $a_i^t = M$, the probability of unmasking is $\Delta t\frac{1 + \eta t}{1 - t}$. If we do unmask, the unmasked state is selected according to our learned model $p_{1|t}^{\theta}(a_1^i | A_t)$. If we are currently in an unmasked state, the probability of switching to the masked state is $\eta \Delta t$.

Hyperparameters and additional sampling techniques that we find important to DFM performance are described in Section S5.


\subsection{Multi-Modal Flow Matching} \label{sec:multi-fm}

We have outlined how we can build flow matching models on the individual modalities in a molecule: continuous flow matching for atomic coordinates and DFM for atom types, charges, and bond orders. Now we describe how to train one model to sample all of these modalities simultaneously. We follow the theoretical formulation of multi-modal flow matching as developed by \citet{campbell_generative_2024}. We can think of the problem of sampling the real distribution of molecules, $p_{data}(g)$, as sampling the joint distribution over modalities, $p_{data}(X,A,C,E)$. The conditional probability path for a molecule is designed to factorize over modalities. That is, modalities are conditionally independent given pairs of initial and final molecular graphs $g_0$ and $g_1$.

\begin{equation} \label{eq:gcondpath}
    p_t(g|g_0,g_1)= \prod_{m \in \{X,A,C,E\}} p_t(m|m_0,m_1)
\end{equation}

At training time, we obtain $g_t$ by sampling the conditional paths of each modality independently. We train one neural network $f_{\theta}$ that takes $g_t$ as input and has separate ``prediction heads'' for each modality $f_\theta(g_t) = \hat{g}_1(g_t) = (\hat{X}_1, \hat{A_1}, \hat{C_1}, \hat{E}_1)$.  The atomic coordinate prediction head $\hat{X}_1 
\in \mathbb{R}^{N \times 3}$ is subjected to the continuous flow matching endpoint loss \eqref{eq:denoiseloss}; it is trained to approximate the final coordinates for the trajectory. The categorical outputs $(\hat{A_1}, \hat{C_1}, \hat{E}_1)$ contain logits over the possible discrete states for each atom/bond; these are subjected to the cross-entropy loss from DFM \eqref{eq:dfm-loss}. The overall loss function for training FlowMol3 is a weighted sum of per-modality flow matching losses:

\begin{equation}
    \mathcal{L} = \sum_{m \in \{X,A,C,E\}} \lambda_m \mathcal{L}_m
\end{equation}

where the losses $\mathcal{L}_m$ are defined as in Algorithm~\ref{alg:train_flowmol}.

After training, the neural network outputs $(\hat{X}_1, \hat{A_1}, \hat{C_1}, \hat{E}_1)$ parameterize a vector field on atomic coordinates and CTMCs on each of the categorical modalities that, when sampled simultaneously, will produce molecules from the target data distribution as $t\to1$. We provide simplified training and sampling procedures in Algorithm \ref{alg:train_flowmol} and Algorithm \ref{alg:sample_flowmol}, respectively.

\begin{algorithm}[H]
\caption{Training FlowMol3}
\label{alg:train_flowmol}
\begin{algorithmic}[1]
\Require Coupling distribution $\pi^*(g_0,g_1)$, loss weights $\{\lambda_m\}_{m\in\{X,A,C,E\}}$, learning rate $\eta$, untrained FlowMol network $f_\theta$
\For{each training step}
    \State Sample a pair of molecular graphs $(g_0, g_1)\sim p(g_0,g_1)$
    \Statex\quad\quad $g_1 \sim p_{data}(g)$
    \Statex\quad\quad $X_0 \sim \mathcal{N}(0,I)$
    \Statex\quad\quad $X_0 \leftarrow Align(X_0,X_1)$ \hfill $\triangleright$ Section S3
    \Statex\quad\quad $A_0 = C_0 = E_0 = \text{mask}$
    \State Sample time $t\sim \mathcal{U}(0,1)$
    \State Sample conditional probability paths:
    \Statex\quad\quad $X_t \leftarrow (1-t)X_0 + t\,X_1$
    \Statex\quad\quad $A_t \sim p_t(A\,|\,A_0,A_1)$ \hfill $\triangleright$ Eq. \eqref{eq:ctmc_cond_path}
    \Statex\quad\quad $C_t \sim p_t(C\,|\,C_0,C_1)$ \hfill $\triangleright$ Eq. \eqref{eq:ctmc_cond_path}
    \Statex\quad\quad $E_t \sim p_t(E\,|\,E_0,E_1)$ \hfill $\triangleright$ Eq. \eqref{eq:ctmc_cond_path}
    \State Let $g_t = (X_t, A_t, C_t, E_t)$
    \State Predict endpoint $f_\theta(g_t) = \hat{g}_1(g_t) = (\hat{X}_1, \hat{A_1}, \hat{C_1}, \hat{E}_1)$
    \State Compute per‐modality losses
    \Statex\quad\quad $\mathcal{L}_X = \mathcal{L}_{EFM}(\hat{X}_1, X_1)$
    \Statex\quad\quad $\mathcal{L}_A = \mathcal{L}_{CE}(\hat{A}_1, A_1)$
    \Statex\quad\quad $\mathcal{L}_C = \mathcal{L}_{CE}(\hat{C}_1, C_1)$
    \Statex\quad\quad $\mathcal{L}_E = \mathcal{L}_{CE}(\hat{E}_1, E_1)$
    \State Total loss: $\mathcal{L} = \sum_{m\in\{X,A,C,E\}}\lambda_m\,\mathcal{L}_m$
    \State Update parameters: $\theta \leftarrow \theta - \eta\,\nabla_\theta \mathcal{L}$
\EndFor
\end{algorithmic}
\end{algorithm}

\begin{algorithm}[H]
\caption{Sampling FlowMol3}
\label{alg:sample_flowmol}
\begin{algorithmic}[1]
\Require Trained model $f_\theta$, number of steps $K$
\State Initialize $g_{0} = (X_{0}, A_{0}, C_{0}, E_{0})$ by sampling:
\Statex\quad $X_{0}\sim\mathcal{N}(0,I)$,\quad $A_{0}=C_{0}=E_{0}=\mathrm{mask}$
\For{$k = 0,\dots,K-1$}
    \State $t = \tfrac{k}{K},\;\Delta t = \tfrac{1}{K}$
    \State Predict endpoint $f_\theta(g_t) = \hat{g}_1(g_t) = (\hat{X}_1, \hat{A_1}, \hat{C_1}, \hat{E}_1)$
    \State \textbf{Continuous update:}
    \Statex\quad $\displaystyle u_{\theta}(X_{t},t) = \frac{\hat X_{1} - X_{t}}{1 - t}$
    \Statex\quad $X_{t+\Delta t} = X_{t} + u_{\theta}(X_{t},t)\,\Delta t$
    \State \textbf{Discrete update:}
    \For{each modality $m\in\{A,C,E\}$ and atom or bond $i$}
        \State Compute $u^{i}_{m}(\,\cdot\mid g_{t})$ \hfill $\triangleright$ Eq. \eqref{eq:marginal_pvel}
        \State Sample $m^i_{t+\Delta t}\sim \mathrm{Cat}\left(\delta(m^i_t, m^i_{t+\Delta t}) + u^{i}_{m}( m^i_{t+\Delta t} | g_{t}) \Delta t \right) $
    \EndFor
    \State $g_{t+\Delta t} \leftarrow \bigl(X_{t+\Delta t},\,A_{t+\Delta t},\,C_{t+\Delta t},\,E_{t+\Delta t}\bigr)$
\EndFor
\State \Return final molecule $g_{1}$
\end{algorithmic}
\end{algorithm}

\subsection{Model Architecture}

FlowMol3 is a message-passing graph neural network (GNN). Molecules are treated as fully-connected, directed graphs where every atom is a node. The GNN operates on graphs where nodes have Cartesian positions $x_i \in \mathbb{R}^3$, scalar features $s_i \in \mathbb{R}^{d_s}$, and vector features $v_i \in \mathbb{R}^{d_v \times 3}$. We also model edges as having scalar features $e_{ij} \in \mathbb{R}^{d_e}$. The operations applied to node positions and vector features are SE(3)-equivariant while the operations on node scalar and edge scalar features are SE(3)-invariant. Node vector features are geometric vectors (vectors with rotation order 1) that are relative to the node position. 

The coordinates of atoms $X$ correspond to node positions. The initial scalar node features are produced by passing atom type, charge, and time embeddings through a shallow MLP. Similarly the initial edge features are obtained by embedding the bond order along each edge. Node vector features are initialized to zeros and are learned through the message-passing routine as functions of the relative positions between atoms. 

\begin{figure}[tb]
    \centering
    \includegraphics[width=0.47\textwidth]{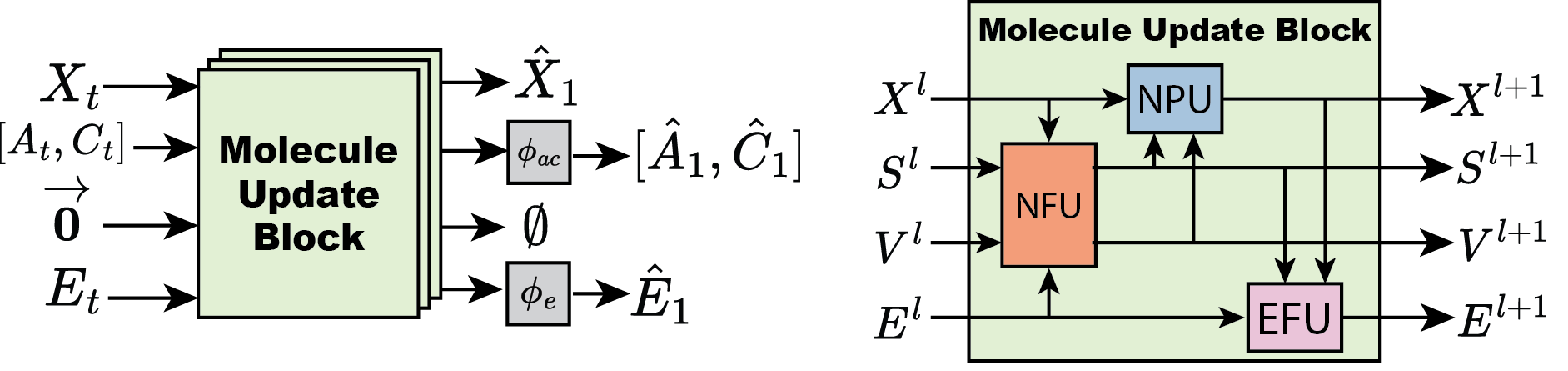}
    \caption{\textbf{FlowMol Architecture} \textit{Left}: An input molecular graph $g_t$ is transformed into a predicted final molecular graph $g_1$ by being passed through multiple molecule update blocks. \textit{Right:} A molecule update block uses NFU, NPU, and EFU sub-components to update all molecular features.}
    \label{fig:arch}
\end{figure}

Graph features are iteratively updated within the neural network architecture by passing through a chain of Molecule Update Blocks (MUB). After passing through MUB layers, the final positions are the predicted final positions of the molecule. The node scalar features are decoded to atom type and charge logits ($\hat{A}_1, \hat{C}_1$) via shallow MLPs. Similarly, edge features are decoded to bond order logits $\hat{E}_1$ via a shallow MLP. The model architecture is visualized in Figure \ref{fig:arch}.

A Molecule Update Block contains three components: a node feature update (NFU), node position update (NPU) and edge feature update (EFU). The NFU performs a message-passing graph convolution to update the scalar and vector features on each node. The NPU and EFU blocks are node and edge-wise operations, respectively. 

Geometric Vector Perceptrons (GVPs) \cite{jing_equivariant_2021} are used to parameterize learnable functions that operate on equivariant features, such as the message-generating functions in graph convolutions. A GVP can be thought of as a single-layer neural network that applies linear and point-wise non-linear transformation to its inputs. The difference between a GVP and a conventional feed-forward neural network is that GVPs operate on two distinct data types: scalar (rotation order 0) and vector (rotation order 1) features. These features exchange information but the operations on scalars are E(3)-invariant and the operations on vector features are E(3)-equivariant. We introduce a variant of GVP that is made SE(3)-equivariant by the addition of cross product operations. FlowMol3 is therefore capable of assigning different likelihoods to stereoisomers. Our cross product variant of GVP is described in Section S6.


\paragraph{Node Feature Update Block} The node feature update block performs a graph convolution to update node scalar and vector features $s_i, v_i$. The message generating and node-update functions for this graph convolution are each chains of GVPs. GVPs accept and return a tuple of scalar and vector features. Therefore, scalar and vector messages $m_{i \to j}^{(s)}$ and $m_{i \to j}^{(v)}$ are generated by a single function $\psi_M$, which is 3 GVPs chained together.

\begin{equation}
    \label{gvp_mij}
    m_{i \to j}^{(s)}, m_{i \to j}^{(v)} = \psi_M \left(
    \left[ s_i^{(l)} : e_{ij}^{(l)} : d_{ij}^{(l)} \right]
    , \left[v_i : \frac{x_i^{(l)} - x_j^{(l)}}{d_{ij}^{(l)}} \right]\right) 
\end{equation}
where $:$ denotes concatenation and $d_{ij}$ is the distance between nodes $i$ and $j$ at molecule update block $l$. In practice, we replace all instances of $d_{ij}$ with a radial basis embedding of that distance before passing through GVPs or MLPs. Message aggregation and node features updates are performed as described in \citet{jing_equivariant_2021}, with the exception that we do not use a dropout layer:

\begin{equation}
    \label{gvp_update}
    s_i^{(l+1)}, v_i^{(l+1)} = \mathrm{LN} \left( [s_i^{(l)}, v_i^{(l)}] + \\
     \psi_U \left( \frac{1}{|\mathcal{N}(i)|} \sum_{j \in \mathcal{N}(i) } \left[ m_{j \to i}^{(s)}, m_{j \to i}^{(v)} \right]   \right) \right) 
\end{equation}
where node update function $\psi_U$ is a chain of three GVPs and $\mathrm{LN}$ is a layer normalization. 

\paragraph{Node Position Update Block} The purpose of this block is to update node positions $x_i$. Node positions are updated:

\begin{equation}
    x_i^{(l+1)} = x_i^{(l)} + \psi_x \left( s_i^{(l+1)}, v_i^{(l+1)} \right)
\end{equation}

Where $\psi_x$ is a chain of 3 GVPs in which the final GVP emits 1 vector and 0 scalar features. The last GVP in $\psi_x$ has its vector-gating activation function ($\sigma_g$ in Algorithm S1) is set to the identity.

\paragraph{Edge Feature Update Block} Edge features, $e$, are updated by the following equation:

\begin{equation}
       e_{ij}^{(l+1)} = \mathrm{LN} \left(e_{ij}^{(l)} + \phi_e \left( s_i^{(l+1)}, s_j^{(l+1)}, d_{ij}^{(l+1)} \right) \right)
\end{equation}

Where $\phi_e$ is a shallow MLP that accepts as input the node scalar features, $s$, of nodes participating in the edge as well as the distance, $d_{ij}$, between the nodes from the positions computed in the NPU block.

\subsection{Enabling Self-Correction}

Based on our experience with FlowMol2,\cite{dunn_mixed_2024} which has a similar architecture to FlowMol3, we developed the hypothesis that transport-based models produce poor-quality samples due to an inability to self-correct and avoid distribution drift. An imperfect denoising model may take the sampling procedure out of distribution. Then, once slightly out of distribution, the performance of the denoising model may further degrade; causing more drift from the target distribution. To address the need for self-correction, we incorporate three additional features into FlowMol3. Through ablation experiments we show that these three features dramatically improve the quality of generated molecules.

\subsubsection{Self-Conditioning}

In vanilla transport-based generative models, the neural network only observes the state of the system at time $t$. Self-conditioning is a technique where the neural network takes its own previous prediction as input in addition to the current state of the system. Originally proposed for text and image generation \cite{chen_analog_2023}, subsequent work has found self-conditioning to substantially boost the performance of molecular generative models \cite{irwin_efficient_2024,reidenbach_applications_2024,yim_fast_2023,stark_harmonic_2024,watson_novo_2023}. 

Self-conditioning is viewed by some as a recycling technique\cite{yim_fast_2023,stark_harmonic_2024}; effectively adding more layers to the network without adding additional parameters. To our knowledge, limited effort has been made to explain why self-conditioning improves performance. We offer the perspective that self-conditioning enables the model to detect and correct inaccurate predictions.  At training time, the model must evaluate its own outputs and determine how to improve them; the model can only improve upon its past predictions by having some ability to find fault in them. 

The denoising neural network takes in a noisy molecule $g_t$ and outputs a predicted final molecule $\hat{g}_1(g_t)$. To implement self-conditioning, we modify our network so that it can \textit{optionally} take a past prediction as an additional input, $\hat{g}_1(g_t,\hat{g}_1(g_{s}))$ where $s=t-\Delta t$. 

At training time, we first predict a denoised molecule using only the current system state $\hat{g}_1(g_t)$. In 50\% of training steps we compute losses and take a gradient step on this prediction. For the other 50\% of training steps, we pass the prediction back through the neural network, $\hat{g}_1(g_t,\hat{g}_1(g_{t}))$, and then compute losses on this quantity. In the latter case, the first pass through the neural network is done without keeping gradients in the computation graph, so the training-time overhead is minimal. 

At inference time, we always pass in the network's prediction from the previous integration step along with the current step; $\hat{g}_1(g_t,\hat{g}_1(g_{s}))$ where $s=t-\Delta t$; this incurs minimal overhead compared to inference without self-conditioning.

Our network is able to ``optionally'' take a past prediction as input because we implement a simple self-conditioning module as a residual layer similar to \citet{reidenbach_applications_2024}. The self-conditioning module produces residual node and edge embeddings that quantify the difference between $g_t$ and $\hat{g}_1(g_s)$. These embeddings are added to the node and edge embeddings of $g_t$ just before passing through the Molecule Update Blocks. 

\subsubsection{Fake Atoms}

Typically, transport-based models for molecules are fixed-dimensional. That is, the number of atoms in the system is chosen before running inference and is unable to change during inference. We introduce an additional atom type, the ``fake atom'' type, that enables the model to dynamically remove and add atoms to the system at inference time.

In transport-based models, the predicted final molecule does not change very much at $t>0.4$ (see Figure \ref{fig:xhat_movement}). From observing $\hat{g}_1$ inference trajectories from FlowMol2, we noticed instances where there were not enough atoms in a region of a molecule to form typical topological structures. In these cases, the atom cannot be moved to a completely different region of the molecule in the limited number of timesteps remaining. The model would instead adjust the local topological configuration to produce functional groups that, while technically valid, were unstable or rare in the data distribution. A functional group analysis showed an over-representation of heteroatom-containing functional groups such as epoxides, peroxides, and two heteroatoms separated by a single carbon (see Figure \ref{fig:reos-flags} and Section S8 in the Supplementary Information). We hypothesized that equipping the model with the ability to adjust the number of atoms would alleviate these issues.

At training time, a random number of ``fake atoms'' is added to the ground-truth molecule $g_1$. The number of fake atoms is sampled from a uniform distribution $\mathcal{U}(0, Np)$ where $N$ is the number of real atoms and $p$ is a hyperparameter we set to $0.3$. Each fake atom is assigned an ``anchor'' atom. The positions of fake atoms are Gaussian offsets from their anchor atoms $x_{\mathrm{fake}} \sim \mathcal{N}(x_{\mathrm{anchor}}, \mathbb{I})$.

At training time, to correctly identify a fake atom, the model must essentially identify that the fake atom position cannot be re-arranged to form a realistic molecule. At inference time, if atoms enter an arrangement that is out-of-distribution, the model may recognize the system state as one where ``fake atoms'' are present, and thus use that mechanism to propose something that is in distribution. Including fake atoms prevents the model from truly seeing out-of-distribution structures at inference time, even if drift occurs, and imparts the model with a mechanism of correction for these instances.

Concurrently to our work, \citet{schneuing_multi-domain_2024} proposed the addition of a removable atom type for receptor-conditioned \textit{de novo} design. Although their implementation differs slightly from ours (different numbers of fake atoms are added to the system, and fake atom positions are placed at the ligand center of mass), they show that this feature improves the quality of designed molecules.

\subsubsection{Late-Stage Geometry Distortion}

At training time, for any graph with $t > 0.5$, the conditional probability path is modified so that additional perturbations are applied to the coordinates of a subset of atoms in the molecule. The conditional path is modified as follows:

\begin{equation} \label{eq:distortion-path}
    X_t = (1 - t)\,X_0 + t\,X_1 + \mathbb{I}[t \geq t_{\mathrm{distort}}]\,\bigl(M \odot \varepsilon\bigr),
\end{equation}
where $\mathbb{I}$ is the indicator function, $\odot$ is the Hadamard product, $M \in [0,1]^N$ is a binary mask over atoms having the property $M_i \sim \mathrm{Bernoulli}(p_{\mathrm{distort}})$, and $\epsilon \in \mathbb{R}^{N\times 3}$ is a per-atom displacement having the property $\epsilon_i \sim \mathcal{N}(0, \sigma_{\mathrm{distort}} I_3)$. Geometry distortion is controlled by three hyperparameters that are set to $p_{\mathrm{distort}} = 0.2$, $t_{\mathrm{distort}} = 0.5$, and $\sigma_{\mathrm{distort}} = 0.5$.

The motivation for this feature is that transport-based models produce suboptimal geometries during inference despite never observing suboptimal geometries at training time. With geometry distortion, the model should be able to propose corrections after distribution drift has occurred in order to bring the system back in-distribution.

If we choose $p_{\mathrm{distort}} = 1.0$ and $t_{\mathrm{distort}} = 0$, then we recover Gaussian conditional probability paths proposed in seminal flow matching works. Theoretical \cite{albergo_stochastic_2023} and empirical \cite{stark_harmonic_2024} arguments have suggested that, when the base distribution is not Gaussian, adding Gaussian noise on top of the interpolants may improve performance by smoothing the learned vector field. SemlaFlow \cite{irwin_efficient_2024} uses a conditional probability path where Gaussian noise is added to all atoms at all times, but presents no ablations to quantify the effect of this. 

We chose only to apply distortion near the end of trajectories (by setting $t_{\mathrm{distort}} = 0.5$) because at earlier times, the samples are closer to samples from the standard Gaussian (the prior), and so adding more Gaussian noise to them would do little to alter the distribution. We proposed the masking of added noise so that the model observes both valid and perturbed geometries; potentially encouraging the model to distinguish between them. 

\section{Experiments}

\subsection{Training Data}

We train models on the GEOM-Drugs dataset \cite{axelrod_geom_2022} using explicit hydrogens. GEOM-Drugs contains drug-like molecules with up to 30 conformers per molecule. Each conformer has undergone a relaxation with respect to the GFN2-xTB semi-empirical quantum chemical potential \cite{bannwarth_gfn2-xtbaccurate_2019} and sits at a local minima of this potential. We use the same dataset splits as \citet{vignac_midi_2023} which, to our knowledge, are purely random splits. The training dataset contains 243,480 unique molecules and 5,741,535 conformers; the mean and median number of conformers per molecule is 23.6 and 30.0, respectively. As suggested in \citet{nikitin_geom-drugs_2025}, we kekulize all molecules in the dataset and do not explicitly model aromatic bonds.

\subsection{Metrics} \label{sec:metrics}

\subsubsection{Validity}

We report ``\% Valid'' which is the percent of molecules that can be sanitized by rdkit. We additionally report ``\% PB-valid,'' which is the percent of molecules that are valid and pass several other physical plausibility checks implemented in the PoseBusters package \cite{buttenschoen_posebusters_2024}. These additional checks include whether the molecule forms one component (not fragmented into separate molecules), has reasonable bond lengths and angles, has planar aromatic rings and planar double bonds, has no steric clashes, and whether the internal energy of the conformer is reasonably close to that of an ensemble of conformers produced by ETKDGv3 \cite{wang_improving_2020} and UFF \cite{rappe_uff_1992} minimization. 

\subsubsection{Functional Group Composition}

We argue that producing ``valid'' molecules is a necessary but insufficient condition for a practically useful molecular generative model. A model should be capable of reproducing similar topological configurations as in the training data (i.e., functional groups). Small-molecule drug discovery scientists have compiled sets of functional groups known to be unstable, toxic, or produce erroneous assay results\cite{walters_recognizing_1999}\footnote{Determining whether a functional group is problematic in the context of a drug discovery campaign is subjective. However, measuring the presence of functional groups can still quantify similarity to training data at higher-order levels of organization than chemical valency.}. Taking inspiration from \citet{walters_generative_2024} we measure the frequency at which these functional groups occur in generated molecules and the training data. We specifically use the well-known Dundee \cite{brenk_lessons_2008} and Glaxo Wellcome \cite{hann_strategic_1999} functional group sets. We report the sum of absolute difference between the frequency of functional groups in the training data and in generated molecules; we refer to this metric as the functional group deviation:

\begin{equation}
    \textrm{FG dev.} = \sum_{f \in \mathcal{F}} | \omega_f^{\textrm{train}} - \omega_f^{\textrm{generated}}|
\end{equation}
where $\mathcal{F}$ is the set of functional groups analyzed, $\omega_f^{\textrm{train}}$ and $\omega_f^{\textrm{generated}}$ are the frequency of functional group $f$ in the training data and generated molecules, respectively. Here we define a functional group frequency as the number of instances of the functional group divided by the number of molecules in the sample.

In addition, we count all of the unique ring systems observed in a batch of molecules. We then record how many times each unique ring system is observed in ChEMBL \cite{zdrazil_chembl_2024}, a database of 2.4M bio-active compounds. We report the rate at which ring systems occur that are never observed in ChEMBL; we refer to this metric as the out-of-distribution (OOD) Ring Rate. Examples of OOD ring systems are provided in Section S7. Ring system and structural alert counting are implemented using the useful\_rdkit\_utils repository \cite{walters_patwaltersuseful_rdkit_utils_2024}. 

\subsubsection{Energy}

We use an energy evaluation that measures how closely the molecules generated adhere to the reference potential of the training dataset, as described in \citet{nikitin_geom-drugs_2025}. Briefly, the GEOM dataset contains conformers that are located at local energy minima of the GFN2-xTB semi-empirical quantum chemical potential \cite{bannwarth_gfn2-xtbaccurate_2019}; therefore, generated molecules should also meet this criteria. We perform an energy minimization on generated molecules with respect to this potential. We measure the change in potential energy resulting from this minimization (denoted $\Delta E_{\mathrm{relax}}$) and the all-atom RMSD between the original and minimized conformation. We report the median of both of these metrics which indicate how closely the generated molecules adhere to the energetic states seen at training time. We use the scripts released with \citet{nikitin_geom-drugs_2025} to perform minimization and record the resulting energy and conformational changes. 

\subsection{Sampling}

To generate molecules from FlowMol3 for analysis, we use evenly-spaced timesteps and Euler integration. All sampling of FlowMol models in this work is done with 250 integration steps.

\subsection{Ablations}

To demonstrate the importance of our self-correcting features, we train four ablated versions of FlowMol3: three variants where one of the three self-correction features is removed, and a fourth variant where all three features are removed. 

\subsection{Comparison to Existing Methods}

We compare FlowMol3 to four diffusion models, MiDi \cite{vignac_midi_2023}, JODO \cite{huang_learning_2023}, EQGAT-Diff \cite{le_navigating_2023}, and Megalodon \cite{reidenbach_applications_2024}, and to two flow matching models, SemlaFlow\cite{irwin_efficient_2024} and ADiT \cite{joshi_all-atom_2025}.

SemlaFlow is the most similar method to FlowMol3; it is also a multi-modal flow matching model with a geometric GNN-based architecture. Megalodon and ADiT opted for the well-tested diffusion-transformer architecture that can be readily scaled to large parameter counts \cite{peebles_scalable_2023}. ADiT also notably discards other components that have become somewhat common: equivariance, multi-modal flows, and explicit bond modeling.

For all baseline models except ADiT we sampled 5000 molecules from the trained models using default settings provided by the authors. For ADiT we used a collection of 10 thousand molecules provided by the authors. Metrics on the sampled molecules were then computed using the same script in the FlowMol repository. Sampled molecules are split randomly into 5 subsets before computing metrics; we obtain 95\% confidence intervals on the mean metric values by assuming the sample mean is normally distributed with the standard deviation in means obtained from the five subsets. 

\section{Results}

\subsection{FlowMol3 Achieves State-of-the-Art Performance}

We compare FlowMol3 to existing state of the art methods in Table \ref{tab:model_comparison}. Molecules produced by FlowMol3 are valid 99.9\% of the time; a significant milestone for all-atom generative models. FlowMol3 achieves a PB-validity rate that is 3.6 percentage points higher than the next-best model (SemlaFlow). Moreover, FlowMol3 pushes the PB-validity rate of generated molecules nearly to that of the training data (92\% vs 93\%). The performance of select models on each individual test run by the PoseBusters suite is included in Section S1. 

FlowMol3 achieves the best performance across all evaluations in Table \ref{tab:model_comparison} with the exception of the median energy change and RMSD from energy minimization. However, FlowMol3 is still practically comparable to the best performing methods on these metrics. The median $\Delta E_{\mathrm{relax}}$ is only $0.6$ kcal/mol higher than that of Megalodon. SemlaFlow produced the lowest median RMSD from relaxation; however, the median $\Delta E_{\mathrm{relax}}$ is still $\approx 8.4$x larger than that of FlowMol3. This implies that SemlaFlow produces small inaccuracies in atom placement that incur a high energetic cost. 

To our knowledge, ADiT \cite{joshi_all-atom_2025} is the only model that achieves a similar validity level to FlowMol3, but its molecules deviate more substantially from training data in terms of the topological (0.4 FG Dev., 0.16 OOD ring rate) and energetic/geometric composition (79 kcal/mol $\Delta E_{\mathrm{relax}}$, 83\% PB-valid).

In addition to producing more valid molecules, FlowMol3 also produces molecules with functional groups that are more similar to its training data than existing methods. This is indicated by a 23\% reduction in functional group deviation over the next best model (SemlaFlow) and a 38\% reduction in the frequency of OOD ring systems over the next best model (ADiT). It is worth noting that despite this progress, the difference in chemistry between generated molecules and ``real'' molecules is still practically significant.

FlowMol3 is also highly parameter efficient; delivering state of the art performance while being substantially smaller than comparable models. The most comparable models in terms of performance - SemlaFlow, Megalodon, and ADiT - have 6.7, 10, and 25 times more learnable parameters, respectively.

\begin{table*}
\caption{FlowMol3 Comparison with Existing Models: Mean value of metrics computed on molecules sampled from each model, along with 95\% confidence on the mean metric values as estimated from 5 repeat samplings. All models were trained on GEOM-Drugs with explicit hydrogens. RMSD is measured in Angstroms. Detailed descriptions of all metrics are available in Section \ref{sec:metrics}. The first row of the table is the corresponding metric value computed on the training data. To estimate \% PB-Valid on the training data, due to computational constraints, we performed 10 repeated random samplings of 15 thousand molecules from a version of the dataset containing 5 conformers per unique molecule; the value presented is the sample mean with a 95\% confidence interval.}
\label{tab:model_comparison}
\resizebox{\textwidth}{!}{%
\begin{tabular}{lrrrrrrr}
\toprule
Model & \% Valid $(\uparrow)$ & \% PB-Valid $(\uparrow)$ & FG Dev. $(\downarrow)$ & OOD Rings $(\downarrow)$ & Med. $\Delta E_{\mathrm{relax}}$ $(\downarrow)$ & Med. Relax RMSD $(\downarrow)$ & Params (M) \\
\midrule
training data & $100.00$ & $93.2 \pm 0.1$ & $0.00$ & $0.05$ & $0.00$ & $0.00$ &  \\
\midrule
FlowMol3 & $\mathbf{99.9\pm 0.1}$  & $\mathbf{91.9\pm 0.7}$  & $\mathbf{0.27\pm 0.03}$  & $\mathbf{0.10\pm 0.01}$  & $3.83\pm 0.08$  & $0.39\pm 0.01$ & $6$ \\
SemlaFlow \cite{irwin_efficient_2024} & $95.5\pm 0.5$ & $88.5\pm 1.3$ & $0.35\pm 0.02$ & $0.20\pm 0.02$ & $31.92\pm 2.30$ & $\mathbf{0.24\pm 0.03}$  & $40$ \\
Megalodon \cite{reidenbach_applications_2024} & $94.8\pm 0.3$ & $86.6\pm 0.7$ & $0.39\pm 0.03$ & $0.17\pm 0.00$ & $\mathbf{3.17\pm 0.11}$  & $0.41\pm 0.01$ & $60$ \\
ADiT\cite{joshi_all-atom_2025} & $99.9\pm 0.0$ & $82.7\pm 0.8$ & $0.41\pm 0.03$ & $0.16\pm 0.01$ & $79.32\pm 1.00$ & $1.30\pm 0.02$ & $150$ \\
EQGAT-Diff\cite{le_navigating_2023} & $86.0\pm 0.9$ & $77.6\pm 0.8$ & $0.58\pm 0.03$ & $0.28\pm 0.01$ & $6.51\pm 0.16$ & $0.60\pm 0.01$ & $12$ \\
JODO \cite{huang_learning_2023} & $78.1\pm 0.9$ & $65.8\pm 0.8$ & $0.43\pm 0.02$ & $0.22\pm 0.00$ & $10.11\pm 0.19$ & $0.73\pm 0.01$ & $6$ \\
Midi \cite{vignac_midi_2023} & $72.9\pm 2.5$ & $59.1\pm 2.1$ & $0.54\pm 0.02$ & $0.33\pm 0.01$ & $19.63\pm 0.65$ & $0.86\pm 0.01$ & $24$ \\
\bottomrule
\end{tabular}
}
\end{table*}

\subsection{Effects of Self-Correction on Molecule Quality}

We introduce three features to FlowMol that we argue enable the model to better perform error detection and correction at inference time: self-conditioning, fake atoms, and geometry distortion. The impact of these features on molecule quality is quantified in Table \ref{tab:model_metrics}.

While removing any one of these features results in a modest performance degradation, removing all three produces a dramatic difference. The data in Table \ref{tab:model_metrics} suggests there are positive interaction effects between these features. For example, removing just one feature causes (+2, -1, -3) percentage point changes to \% PB-Valid but removing all 3 features causes a 14 point reduction.

A potentially useful analysis of the single-feature removal ablations is to note which of the three features had the most significant impact on each of the metrics evaluated. Chemical composition-related metrics (FG dev. and OOD ring rate) were most impacted by removing fake atoms. The energy of relaxation was increased the most by removing distortion. Validity and posebusters validity were most impacted by self-conditioning. 

Fake atoms seem to have a strong positive impact on the chemistry/topology of generated molecules but a less substantial impact on the validity and relaxation energies. In contrast, geometry distortion has the most substantial impact on the relaxation energies but a smaller effect on the chemistry/topology. An interpretation is that these features are each serving similar functions (imparting robustness to distribution drift) but the former is more impactful to discrete modalities and the latter to continuous modalities. 


\begin{table*}
\caption{
Ablations of Self-Correcting Features: Mean value of metrics computed on molecules sampled from ablations of the FlowMol3 model, along with 95\% confidence on the mean metric values as estimated from 5 repeat samplings. Each ablation differs only by whether each of the three self-correcting features is used in the model; this information is indicated by the first three columns. All models were trained on GEOM-Drugs with explicit hydrogens. Detailed descriptions of all metrics are available in Section~\protect\ref{sec:metrics}. }
\label{tab:model_metrics}
\resizebox{\textwidth}{!}{%
\begin{tabular}{lllllllll}
\toprule
Self-Cond. & Fake Atoms & Distortion & \% Valid $(\uparrow)$ & \% PB-Valid $(\uparrow)$ & FG Dev. $(\downarrow)$ & OOD Rings $(\downarrow)$ & Med. $\Delta E_{\mathrm{relax}}$ $(\downarrow)$ \\
\midrule
$\checkmark$ & $\checkmark$ & $\checkmark$ & $\mathbf{99.9\pm 0.1}$  & $91.9\pm 0.7$ & $\mathbf{0.27\pm 0.03}$  & $\mathbf{0.10\pm 0.01}$  & $\mathbf{3.83\pm 0.08}$  \\
$\times$ & $\checkmark$ & $\checkmark$ & $99.1\pm 0.2$ & $89.3\pm 0.8$ & $0.29\pm 0.03$ & $0.18\pm 0.01$ & $5.95\pm 0.12$ \\
$\checkmark$ & $\times$ & $\checkmark$ & $99.9\pm 0.0$ & $\mathbf{94.0\pm 0.5}$  & $0.37\pm 0.01$ & $0.23\pm 0.01$ & $4.82\pm 0.11$ \\
$\checkmark$ & $\checkmark$ & $\times$ & $98.6\pm 0.2$ & $90.6\pm 0.3$ & $0.33\pm 0.02$ & $0.20\pm 0.01$ & $6.40\pm 0.09$ \\
$\times$ & $\times$ & $\times$ & $95.1\pm 0.4$ & $78.1\pm 1.1$ & $0.91\pm 0.04$ & $0.32\pm 0.01$ & $14.75\pm 0.34$ \\
\bottomrule
\end{tabular}
}
\end{table*}

\subsection{Mechanistic Interpretation of Feature Effects}

\begin{figure}
    \centering
    \includegraphics[width=0.4\textwidth]{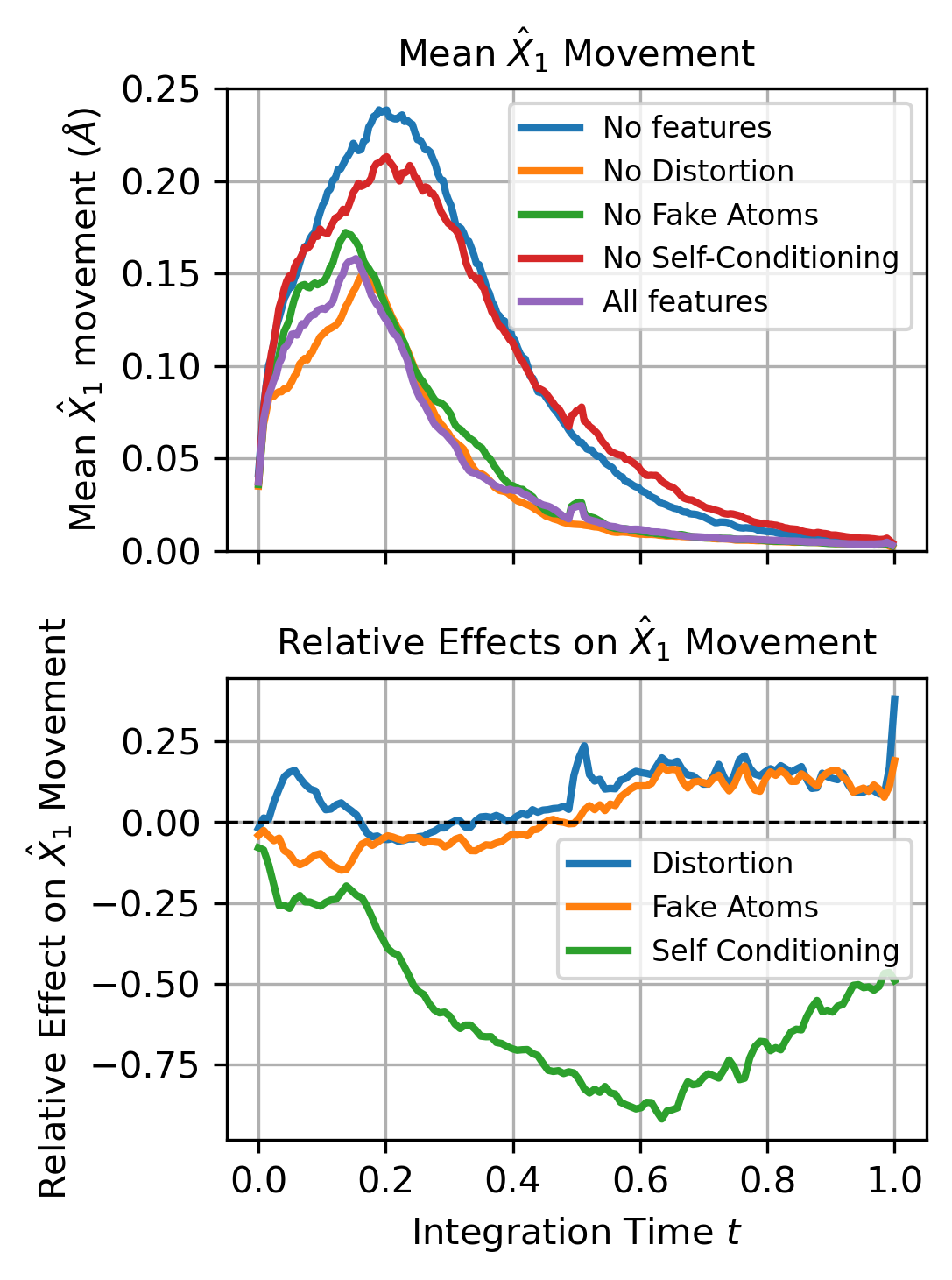}
    \caption{Analysis of Denoiser Trajectories under Ablations: (Top) Mean $\hat{X}_1$ movement over integration time $t$ for ablated versions of FlowMol3; each model distinguished by whether it has self-conditioning, fake atoms, or geometry distortion. (Bottom) The relative effect of each of the three features. At every time step, we fit a linear additive effects model to predict $\hat{X}_1$ movement given whether each feature is present in the model as binary variables. The ``effect'' of a feature is its coefficient under the linear model. What is plotted here is the relative effect: the fold-change in $\hat{X}_1$ movement induced by that feature. A relative effect of 0.2 would mean that including the feature causes a 20\% increase in $\hat{X}_1$ movement.}
    \label{fig:xhat_movement}
\end{figure}

To provide a mechanistic explanation of the effects of our self-correction enabling features, we analyze trajectories of denoiser outputs $\left\{ \hat{g}_1(g_t)\right\}_{t \in [0,1] } $. After obtaining such a trajectory, we compute the change in each atom's predicted final coordinates between sequential integration steps: $\norm{ \hat{X}^i_1(g_{t+\Delta t}) - \hat{X}^i_1(g_{t}) }$. We refer to this quantity as $\hat{X}_1$ movement. $\hat{X}_1$ movement can be interpreted as how much the denoiser is updating its estimated endpoint throughout the trajectory. We average this quantity over atoms and across 100 sampled trajectories. Mean $\hat{X}_1$ movement trajectories for each of the ablated version of FlowMol3 are shown in Figure \ref{fig:xhat_movement}.

We then fit a linear additive effect model on $\hat{X}_1$ movement as a function of whether or not the three ablated features are present in the model; this allows us to isolate the effect of self-conditioning, fake atoms, and geometry distortion on $\hat{X}_1$ movement as a function of integration time by inspecting the coefficients of the additive effects model. The results are shown in Figure \ref{fig:xhat_movement}.

Self-conditioning substantially reduces $\hat{X}_1$ motion throughout trajectories. This suggests that atoms move more directly towards their final state. In each forward pass, the model is making more accurate estimates of the final molecule.  

Under the linear additive effect model, it appears that both fake atoms and geometry distortion causes a 15-20\% increase $\hat{X}_1$ motion late in the trajectories; this can be interpreted as the model making more updates/corrections to nearly complete molecules.

Interestingly, the bump in $\hat{X}_1$ motion over the $t>0.5$ regime due to fake atoms and geometry distortion is more substantial when self-conditioning is not present than when it is. An interpretation is that geometry distortion and fake atoms do indeed enable late-stage corrections but the system is simply less likely to move out of distribution in the first place when self-conditioning is implemented.

\subsection{Functional Group Analysis}

\begin{figure}
    \centering
    \includegraphics[width=\linewidth]{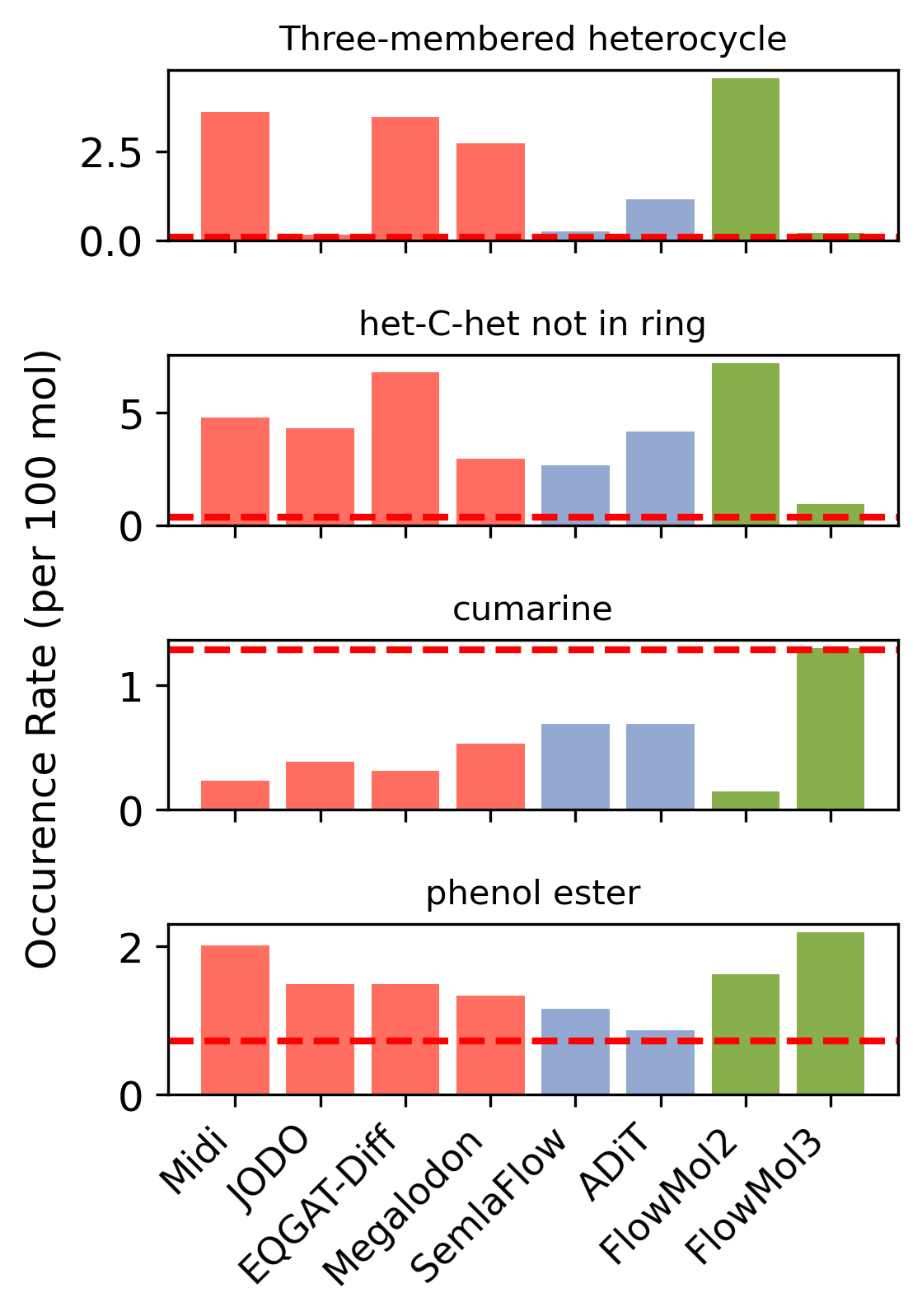}
    \caption{Functional Group Prevalence: Each plot shows the prevalence of a specific functional group among molecules generated by several models. The y-axis is the number of occurrences of the functional group that occurs per 100 molecules sampled. The horizontal red lines on each plot are the frequency of that functional group in the training data. Red bars are for diffusion model baselines, purple bars are for flow matching baselines, and green bars are for FlowMol.}
    \label{fig:reos-flags}
\end{figure}

We find that visualizing the prevalence of individual functional groups in generated samples relative to their prevalence in the training data reveals interesting pathologies. We include a few notable examples in Figure \ref{fig:reos-flags} as well as all functional groups analyzed in Section S8.

Three-membered heterocycles (such as epoxides) are rare in the training data (one instance per 1220 molecules) as they are generally considered a reactive/unstable functional group. Our analysis reveals that this functional group is produced relatively frequently by some of the models evaluated. Our data suggest self-conditioning may dramatically reduce the over-representation of three-membered heterocycles. MiDi, EQGAT-Diff, and FlowMol2, which do not use self-conditioning, produce three-membered heterocycles 44x, 42x, and 56x more frequently than the training data. All but one of the models using self-conditioning produce significantly fewer instances of these functional groups. FlowMol3, JODO, and SemlaFlow produce three-membered heterocycles at 2-3x the rate of the training data. The exception to this trend is Megalodon which implements self-conditioning yet produces three-membered heterocycles 33x more frequently than the training data.

Another functional group that is relatively rare due to instability is the occurrence of two heteroatoms separated by a carbon outside of a ring. This functional group occurs in the training data once per 289 molecules. The evaluated models produce this functional group at 8-20x the frequency of the training data except for FlowMol3, which comes in at 3x the frequency of the training data. This demonstrates that there are features of FlowMol3 that uniquely enable it to match this functional group representation in a way existing models have not been able to, and that this must necessarily be caused by fake atoms or geometry distortion.

Cumarine is another such example of a functional group that FlowMol3 is uniquely capable of recapitulating due to geometry distortion and fake atoms.

In contrast, there are functional groups whose representation are made worse by the addition of self-correcting features; phenol esters are an example of this. There is obviously still room for improvement on the front of matching the chemistry of generated molecules to that of the training data. 

\section{Discussion}

FlowMol3 demonstrates state-of-the-art performance across a broad range of molecular quality metrics. These improvements arise largely from the addition of three techniques: self-conditioning, fake atoms, and late-stage geometry distortion, which we hypothesize mitigate inference-time distribution drift in transport-based generative models.

Beyond increasing the validity of generated molecules, these features have measurable effects on molecular geometry, energetic states, and functional group composition. Their impact is particularly notable given that they do not alter model size or training cost. This suggests that they alleviate an existing pathology that prevented the model from making use of its available computational power.

Our analysis of denoiser trajectories supports this interpretation. Self-conditioning reduces the magnitude of atomic updates during sampling, implying that the model converges to its final predictions more directly. Geometry distortion and fake atoms, meanwhile, appear to improve the model’s ability to maintain or recover from off-distribution states. Together, these features promote sampling dynamics that stay much closer to the desired marginal process.

Despite these advances, FlowMol3 does not eliminate all gaps between generated and real molecules. Functional group composition, in particular, remains imperfectly matched, and certain classes of functional groups continue to be over- or under-represented. While FlowMol3 corrects some of these discrepancies compared to prior models, others persist or are even exacerbated. 

Interestingly, FlowMol3 achieves its performance with substantially fewer parameters than ADiT and other large transformer-based models. This suggests that architectural scale alone may not be sufficient to address the specific pathologies of transport-based generative modeling. Careful study of and improvement in the underlying generative modeling framework is likely necessary as well.

Looking ahead, future research should aim to (1) develop a deeper theoretical understanding of distribution drift in transport-based models, (2) explore whether the self-correction paradigm can be formalized and incorporated at the level of the generative modeling framework itself, and (3) evaluate whether these mechanisms extend to conditional generation tasks.

\section*{Conclusions}

We have presented FlowMol3, a flow-matching-based generative model for 3D \textit{de novo} molecule generation that incorporates three design features—self-conditioning, fake atoms, and late-stage geometry distortion—aimed at improving robustness to inference-time distribution drift. These additions significantly improve the geometric, energetic, and topological quality of generated molecules without increasing model size or training cost.

Our findings suggest that the limitations of prior transport-based generative models may stem less from insufficient model capacity and more from an inability to recover from distribution drift. By addressing this issue directly, FlowMol3 achieves performance that approaches the quality of the training distribution across widely used metrics.

Because the proposed features are architecture-agnostic and inexpensive to implement, they may be readily transferable to other models. More broadly, the principle of designing transport-based generative models to explicitly resist or recover from distribution drift may provide a general strategy for improving the reliability of transport-based generative modeling in molecular design and beyond.

All code, trained models, and evaluation scripts are available at at \url{https://github.com/dunni3/FlowMol} to facilitate reproducibility and to support further work in this area.

\section*{Author contributions}
Ian Dunn conceptualized the work presented, performed investigations, developed methodology, wrote software, wrote the original draft of the manuscript as well as reviewed and edited the manuscript.

David Koes conceptualized the work presented, oversaw the computational experiments and methodology and software development, wrote, reviewed and edited multiple drafts of the manuscript.

\section*{Conflicts of interest}
There are no conflicts to declare.

\section*{Data availability}

All data, code, and trained models for reproducing this work are made publicly available at \url{https://github.com/dunni3/FlowMol}.

\section*{Acknowledgements}

The authors would like to thank Filipp Nikitin, Gabriella Gerlach, and Rishal Aggarwal for helpful discussions in the development of this work.

This work is funded through R35GM140753 from the National Institute of General Medical Sciences. The content is solely the responsibility of the authors and does not necessarily represent the official views of the National Institute of General Medical Sciences or the National Institutes of Health.



\balance


\bibliography{simplexflow} 
\bibliographystyle{rsc} 



\section*{Supplementary material}

# FlowMol3 Supplementary Information

# Contents

# S1   Additional Data on PoseBusters Validity

In the main paper we report the percent of molecules that are valid as determined by the PoseBusters suite[1] under the name "% PB-Valid". A model is considered PB-valid if it passes all of a series of tests. Here we present the pass-rate for each individual test run by PoseBusters; these results are shown in Table S1.

1

Table S1: Pass rates on individual PoseBusters tests with 95% confidence intervals

|  | FlowMol3 | SemlaFlow | Megalodon | ADiT |
|---|---|---|---|---|
| mol pred loaded | $100.0 \pm 0.0$ | $100.0 \pm 0.0$ | $100.0 \pm 0.0$ | $100.0 \pm 0.0$ |
| sanitization | $99.9 \pm 0.1$ | $95.4 \pm 0.5$ | $94.7 \pm 0.2$ | $99.9 \pm 0.0$ |
| inchi convertible | $99.9 \pm 0.1$ | $95.4 \pm 0.5$ | $94.7 \pm 0.3$ | $99.9 \pm 0.0$ |
| all atoms connected | $98.9 \pm 0.2$ | $97.3 \pm 1.1$ | $97.8 \pm 0.5$ | $94.6 \pm 0.5$ |
| bond lengths | $100.0 \pm 0.0$ | $99.6 \pm 0.2$ | $99.9 \pm 0.1$ | $96.0 \pm 0.3$ |
| bond angles | $100.0 \pm 0.0$ | $100.0 \pm 0.0$ | $100.0 \pm 0.0$ | $95.4 \pm 0.6$ |
| internal steric clash | $94.1 \pm 0.5$ | $96.7 \pm 0.6$ | $94.4 \pm 1.1$ | $92.2 \pm 0.4$ |
| aromatic ring flatness | $100.0 \pm 0.0$ | $100.0 \pm 0.0$ | $100.0 \pm 0.0$ | $100.0 \pm 0.0$ |
| non-aromatic ring non-flatness | $99.6 \pm 0.1$ | $99.7 \pm 0.1$ | $99.8 \pm 0.2$ | $97.3 \pm 0.3$ |
| double bond flatness | $99.3 \pm 0.3$ | $99.2 \pm 0.2$ | $99.4 \pm 0.2$ | $99.9 \pm 0.1$ |
| internal energy | $100.0 \pm 0.0$ | $99.9 \pm 0.1$ | $99.9 \pm 0.1$ | $94.4 \pm 0.3$ |
| all | $91.9 \pm 0.7$ | $88.5 \pm 1.3$ | $86.6 \pm 0.7$ | $82.7 \pm 0.8$ |

## S2 Example Molecules

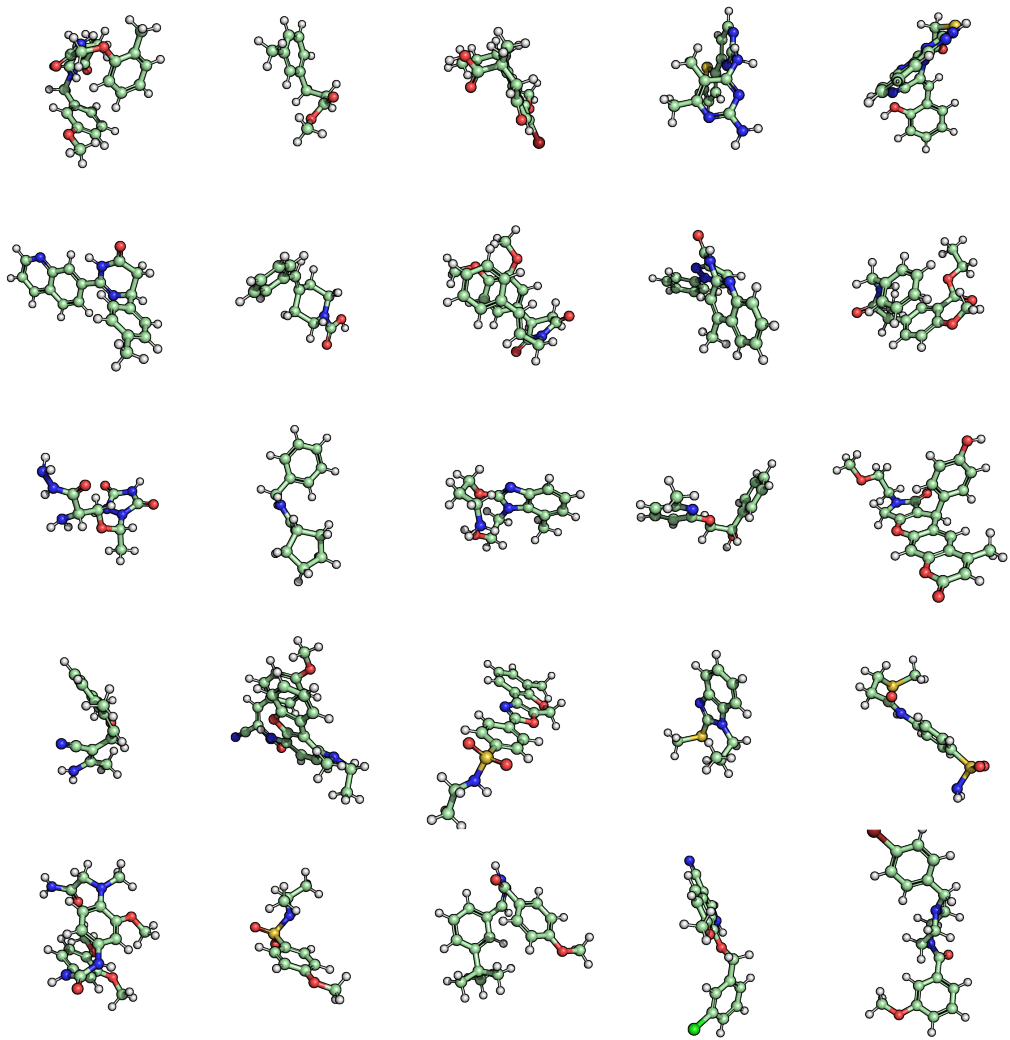

Figure S1: Example molecules sampled from FlowMol3

## S3 The Coupling Distribution

Since we chose the conditioning variable for our conditional probability paths to be pairs of initial and final molecules $(g_0, g_1)$, we must also define a method

3

of obtaining such paired samples of initial and final molecules. In other words, we must define the distribution from which our conditioning variable is obtained $p(g_0, g_1)$. This is generally referred to as a coupling distribution[2–4].

The simplest choice of the coupling distribution would be an independent coupling:

$$p(g_0, g_1) = p(g_0)p(g_1) \qquad (1)$$

In FlowMol, the coupling distribution factorizes over modalities.

$$p(g_0, g_1) = p(X_0, X_1)p(A_0, A_1)p(C_0, C_1)p(E_0, E_1) \qquad (2)$$

All discrete modalities are given the independent coupling:

$$p(A_0, A_1) = p(A_0)p(A_1) \qquad (3)$$

That is, the the target value ($A_1$) is obtained from the dataset, and the corresponding prior value ($A_0$) is always a sequence of mask tokens, independent of the target value.

However, in continuous flow matching, the independent coupling can introduce pathologies that impair model performance such as having a large transport cost (distances between prior and target samples), and intersecting conditional paths. Several works identify this pathology and propose various forms of coupling distributions that involve drawing independent samples and then aligning them in some fashion[3,5,6].

For continuous modalities, we implement a technique similar to the Equivariant Optimal Transport Coupling proposed by Klein *et al.*[6]. We first obtain the $t = 0$ atom coordinates as independent samples from a standard Gaussian $X_0 \sim \prod_{i=1}^{N} \mathcal{N}(x_0^i|0, \mathbb{I}_3)$. We then align the prior coordinates $X_0$ to the ground-truth coordinates $X_1$ via the Kabsch algorithm. Next, we permute the order of atoms in $X_0$. This can be thought of as permuting the intial positions of each atom. We specifically choose the permutation that minimizes the sum of distances between initial and final positions across all atoms (this is known as solving the assignment problem).

## S4  Training Details and Hyperparameters

FlowMol3 is trained with 6 Molecule Update Blocks. Atoms contain 256 hidden scalar features and 32 hidden vector features. Edges contain 128

hidden features. All models are trained for 20 epochs. GEOM models are trained on 4xL40 GPUs with an adaptive batch-size that typically yields 15-25 graphs per batch per GPU. Training takes approximately 4-5 days.

As described in Section 2.4 of the main paper, the overall loss for training FlowMol3 is a weighted sum of per-modality flow matching losses. The loss weights $(\lambda_X, \lambda_A, \lambda_C, \lambda_E)$ are set to (3, 0.4, 1, 2).

All model hyperparameters are visible in the config files provided in our github repository.

# S5 Additional Techniques for DFM Sampling

Here we discuss three techniques we apply to sampling discrete modalities in FlowMol3: remasking, low-temperature sampling, and purity sampling.

## S5.1 Remasking via Stochasticity

Recall our marginal probability velocity ((12) from the main paper, reproduced here)

$$u^i(j, A_t) = \frac{1 + \eta t}{1 - t} p^\theta_{1|t}(a^i_1 = j | A_t) \delta_M(a^i_t) + \eta(1 - \delta_M(a^i_t))\delta_M(j) \quad (4)$$

Includes the hyperparameter $\eta \geq 0$ which can be chosen at inference time. $\eta$ can be called the "stochasticity parameter". If we set $\eta = 0$ then once a sequence element (i.e., an atom type or a bond order) is unmasked, that value is fixed for the remainder of the trajectory. If we choose $\eta > 0$, sequence elements may be remasked and unmasked again repeatedly throughout a trajectory; enabling the denoiser to correct or change past decisions. We find that enabling remasking significantly enhances sample quality. All FlowMol results obtained in this paper are done so using $\eta = 30$.

## S5.2 Low-Temperature Sampling

Recall that for discrete modalities our neural networks directly approximate the distribution of states at $t = 1$ for each sequence element. In the context of atom types, for example, this quantity is denoted $p^\theta_{1|t}(a^i_1 | A_t)$.

We re-normalize logits obtained from the model using a fixed temperature $\tau$:

5

$$p_{1|t}^\theta(a_1^i|A_t) = \text{softmax}\left(\tau^{-1} \log p_{1|t}^\theta(a_1^i|A_t)\right) \quad (5)$$

We find that low-temperature sampling, which biases the sampled discrete states towards the most confident ones, to be critical for model performance. In practice we use $\tau = 0.05$.

### S5.3 Purity Sampling

Under the base DFM formulation, at each integration step, the probability of unmasking each currently-masked sequence element is $\Delta t \frac{1+\eta t}{1-t}$. Every masked token has an equal probability of getting unmasked.

Rather than doing this, we implement purity sampling as described in Campbell *et al.*[7]. We select which currently-masked elements to unmask using a proxy for model confidence. The proxy used for model confidence is the "purity": the maximum category probability output by the model.

During inference, at each integration step, we first sample the number of elements to unmask from a binomial distribution with $n$ equal to the number of mask tokens in the sequence and $p = \Delta t \frac{1+\eta t}{1-t}$. Then after we obtain $k \sim \text{Binomial}(n, p)$, we select the top-$k$ sequence elements with the highest "purity"; these are the tokens that will be unmasked.

## S6 GVP with Cross Product

A geometric vector perception (GVP) can be thought of as a single-layer neural network that applies linear and point-wise non-linear transformation to its inputs. The difference between a GVP and a conventional feed-forward neural network is that GVPs operate on two distinct data types: scalars and vectors. GVPs also allow these data types to exchange information while preserving equivariance of the output vectors. The original GVP only applied linear transformations to the vector features and as a result produces output vectors that are E(3)-equivariant.

We introduce a modification to the GVP as its presented in Jing *et al.*[8]; specifically we perform a cross product operation on the input vectors. The motivation for this is that the cross product is *not* equivariant to reflections. We refer the reader to Appendix F of Schneuing *et al.*[9] for a detailed discussion of the equivariance of cross products. As a result, the version of GVP we present here is SE(3) equivariant. The benefit of being SE(3) equivariant

6

rather than E(3) equivariant is that the generative model becomes sensitive to chiral centers in molecules, since reflecting the molecule will produce different model outputs. The operations for our cross product enhanced GVP are described in Algorithm 1.

---
**Algorithm 1** Geometric Vector Perceptron with Cross Product
---

**Input:** Scalar and vector features: $(s, v) \in \mathbb{R}^f \times \mathbb{R}^{\nu \times 3}$
**Output:** Scalar and vector features: $(s', v') \in \mathbb{R}^j \times \mathbb{R}^{\mu \times 3}$
**Hyperparameter:** Number of hidden vector features $n_h \in \mathbb{Z}^+$
**Hyperparameter:** Number of cross product features $n_{cp} \in \mathbb{Z}^+$

$\quad v_h \leftarrow W_h v \quad \in \mathbb{R}^{n_h \times 3}$
$\quad v_{cp} \leftarrow W_{cp} v \quad \in \mathbb{R}^{2n_{cp} \times 3}$
$\quad v_{cp} \leftarrow v_{cp}[:n_{cp}] \times v_{cp}[n_{cp}:] \quad \in \mathbb{R}^{n_{cp} \times 3}$ // cross product
$\quad v_{h+cp} \leftarrow \text{Concat}(v_h, v_{cp}) \quad \in \mathbb{R}^{(n_h + n_{cp}) \times 3}$
$\quad v_\mu \leftarrow W_\mu v_{h+cp} \quad \in \mathbb{R}^{\mu \times 3}$
$\quad s_{h+cp} \leftarrow \|v_{h+cp}\| \quad \in \mathbb{R}^{n_h + n_{cp}}$
$\quad s_{f+h+cp} \leftarrow \text{Concat}(s, s_{h+cp})$'
$\quad s_j \leftarrow W_j s_{f+h+cp} + b_j \quad \in \mathbb{R}^j$
$\quad s' \leftarrow \sigma(s_j) \quad \in \mathbb{R}^j$
$\quad v' \leftarrow \sigma_g \left( W_g [\sigma^+(s_m)] + b_g \right) \odot v_\mu \text{ (row-wise)} \quad \in \mathbb{R}^{\mu \times 3}$
**return** $(s', v')$

---

# S7 Examples of Out-of-Distribution Ring Systems

To provide intuition about the purpose or significance of the OOD ring system metric, we provide examples of OOD ring systems produced by evaluated models in Figure S2.

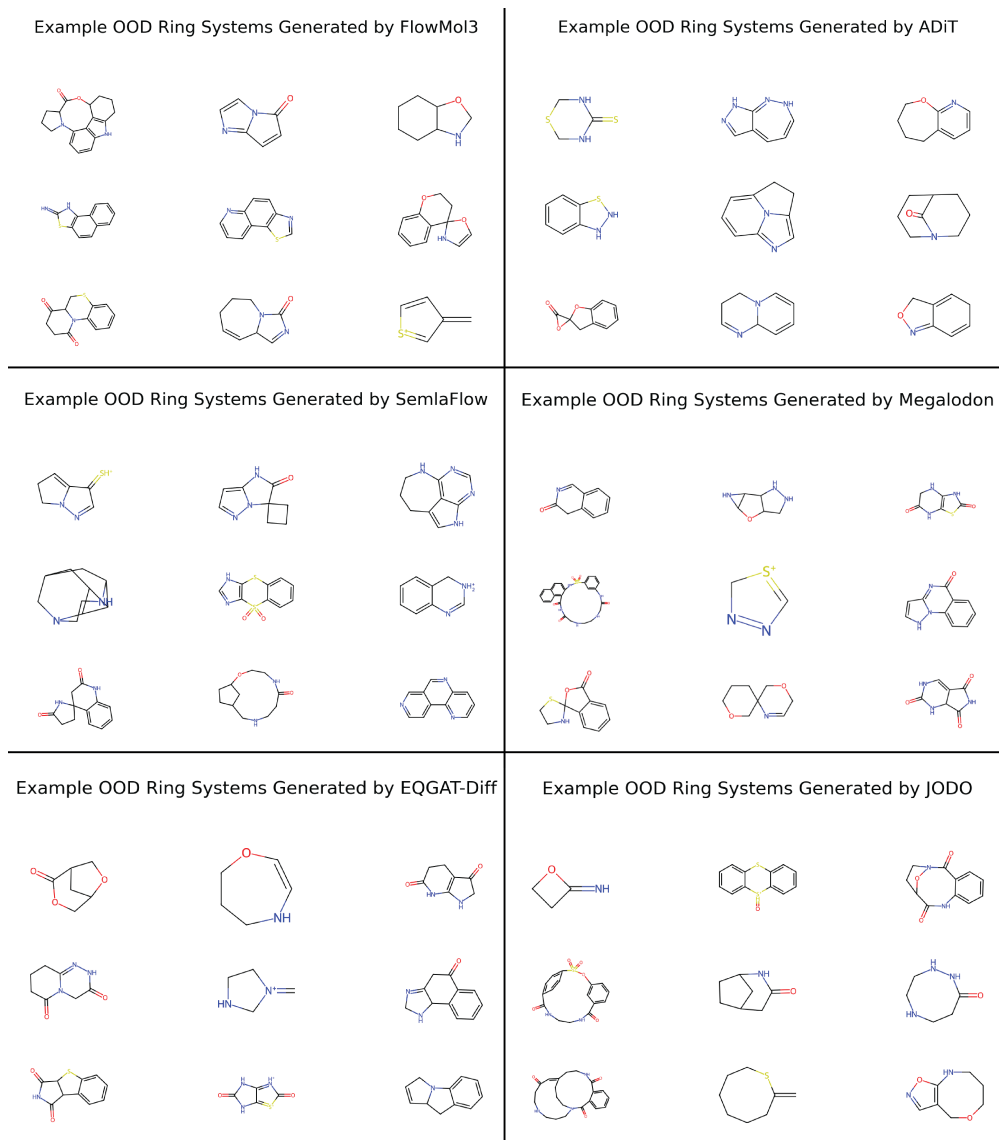

Figure S2: Examples of OOD ring systems (ring systems that do not appear in ChEMBL) that were produced by several of the models evalauted in this paper. The specific OOD ring systems were selected randomly from the set of all OOD ring systems found.

# S8 Individual Functional Group Frequencies

We show the functional group frequencies for the 36 most commonly occuring functional groups evaluated in the following figures. The y-axis in all figures is the frequency of the functional group per 100 sampled molecules.

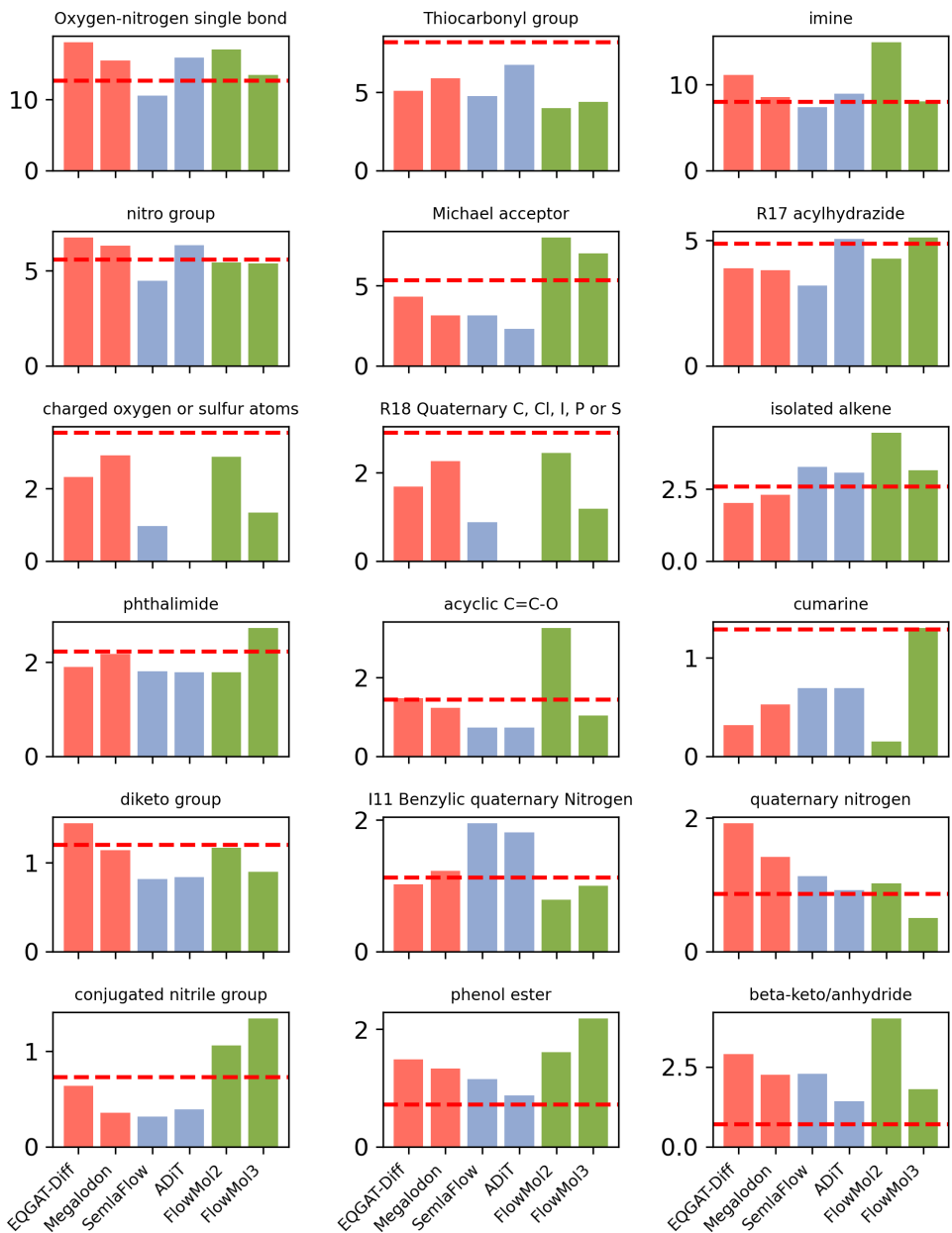

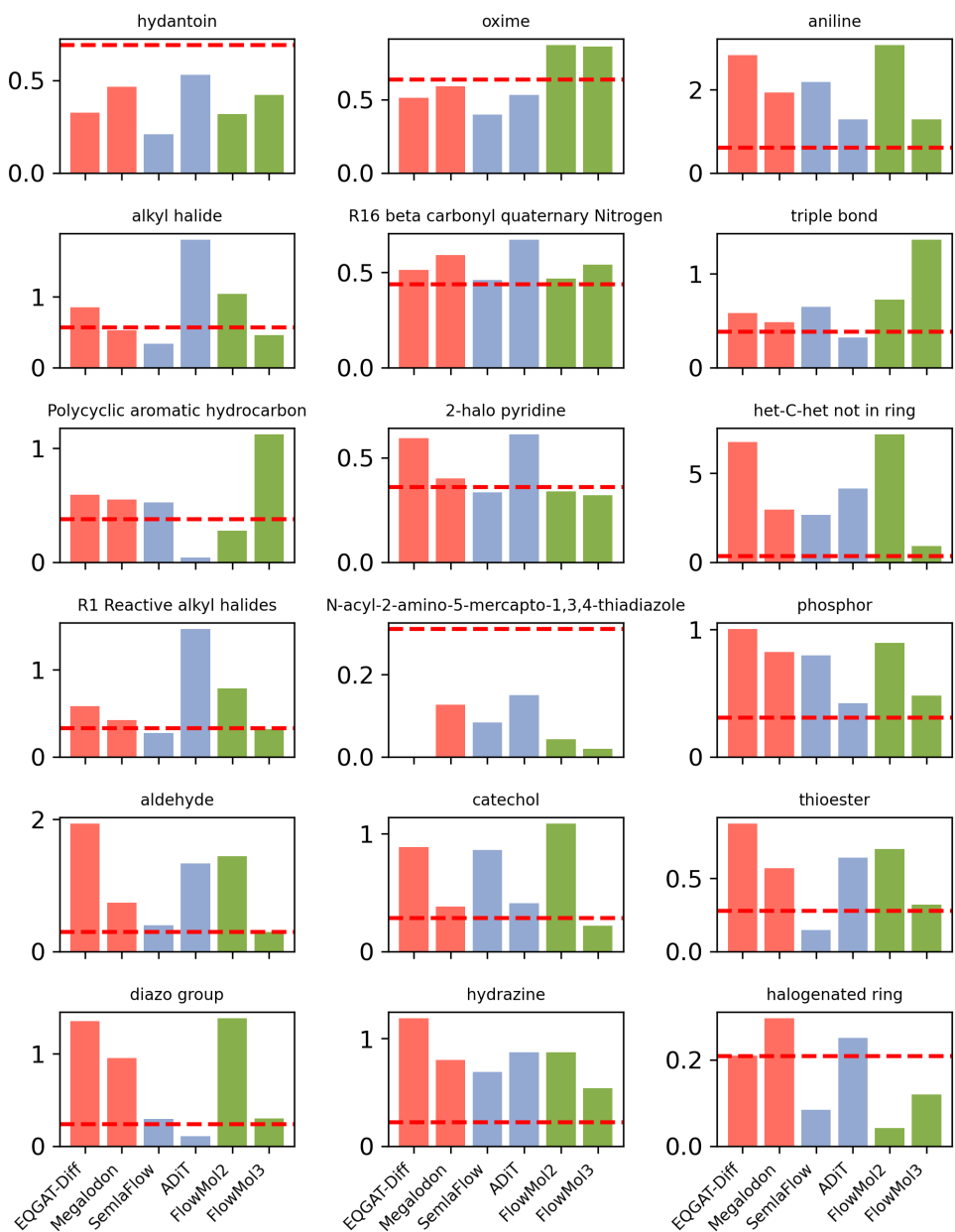



\end{document}